\documentclass[10pt,twocolumn,letterpaper]{article}

\usepackage{titling}

\usepackage{iccv}
\usepackage{times}
\usepackage{epsfig}
\usepackage{graphicx}
\usepackage{amsmath}
\usepackage{amssymb}
\usepackage{makecell}
\usepackage{algorithm}
\usepackage[noend]{algpseudocode}
\usepackage{wrapfig,booktabs}
\usepackage[numbers,sort&compress]{natbib}
\usepackage{color}
\usepackage[normalem]{ulem}
\usepackage[table]{xcolor}
\usepackage{pifont}
\usepackage{multirow}
\usepackage[title]{appendix}
\makeatletter
\@namedef{ver@everyshi.sty}{}
\makeatother
\usepackage{pgfplots}
\pgfplotsset{width=70mm,compat=newest}
\newcommand{\red}[1]{\textcolor{red}{#1}}

\newcommand{\green}[1]{\textcolor{green}{#1}}

\usepackage[pagebackref=true,breaklinks=true,letterpaper=true,colorlinks,bookmarks=false]{hyperref}

\iccvfinalcopy % *** Uncomment this line for the final submission

\def\iccvPaperID{1215} % *** Enter the ICCV Paper ID here

\begin{document}

%%%%%%%%% TITLE
%\title{Reducing Visual Confusion with Discriminative Attention}
%\title{Visual Confusion Reduction with Attention Separation and Consistency}
%\title{Sharpen Focus: Learning Attention with Inter-class Separability}
%\title{Sharpen Focus: Learning Attention with Inter-class Separability and Intra-class Consistency}
\title{Sharpen Focus: Learning with Attention Separability and Consistency}
%\title{Sharpen Focus: Learning Consistent and Separable Attention }
%\title{Learn a Better Visual Explanation: Attention-Guided Training for Image Classification with Visual Confusion Reducing}

%\author{Lezi Wang, Bo Liu\\
%Rutgers University, New Brunswick, NJ\\
%{\tt\small \{lw462, lb507\}@cs.rutgers.edu} %kfliubo@gmail.com}
%% omit the following lines up until the closing ``}''.
%% Additional authors and addresses can be added with ``\and'',
%% just like the second author.
%% To save space, use either the email address or home page, not both
%\and
%Ziyan Wu, Srikrishna Karanam\\
%Kuan-Chuan Peng, Rajat Vikram Singh\\
%Siemens Corporate Technology, Princeton, NJ\\
%{\tt\small \{first.last, singh.rajat\}@siemens.com}
%}

\author{Lezi Wang$^{1}$, Ziyan Wu$^{2}$, Srikrishna Karanam$^{2}$, Kuan-Chuan Peng$^{2}$,\\ Rajat Vikram Singh$^{2}$, Bo Liu$^{1}$, and Dimitris N. Metaxas$^{1}$\\
$^{1}$Rutgers University, New Brunswick NJ\\
$^{2}$Siemens Corporate Technology, Princeton NJ\\
%$^{3}$JD Digits, Mountain View, CA\\
{ \small \{lw462, lb507, dnm\}@cs.rutgers.edu,}
{ \small \{ziyan.wu, srikrishna.karanam, kuanchuan.peng, singh.rajat\}@siemens.com}
}

\maketitle
%\ificcvfinal\thispagestyle{empty}\fi

%%%%%%%%% ABSTRACT
\begin{abstract}

Recent developments in gradient-based attention modeling have seen attention maps emerge as a powerful tool for interpreting convolutional neural networks. Despite good localization for an individual class of interest, these techniques produce attention maps with substantially overlapping responses among different classes, leading to the problem of visual confusion and the need for discriminative attention. In this paper, we address this problem by means of a new framework that makes class-discriminative attention a principled part of the learning process. Our key innovations include new learning objectives for attention separability and cross-layer consistency, which result in improved attention discriminability and reduced visual confusion. Extensive experiments on image classification benchmarks show the effectiveness of our approach in terms of improved classification accuracy, including CIFAR-100 (+3.33\%), Caltech-256 (+1.64\%), ILSVRC2012 (+0.92\%), CUB-200-2011 (+4.8\%) and PASCAL VOC2012 (+5.73\%).

\end{abstract}

%%%%%%%%% BODY TEXT
\section{Introduction}
Visual recognition has seen tremendous progress in the last few years, driven by recent advances in convolutional neural networks (CNNs) \cite{he2016deep,jetley2018learnICLR}. Understanding their predictions can help interpret models and provide cues to design improved algorithms.

\begin{figure}[t]
    \centering
    \includegraphics[width=\columnwidth]{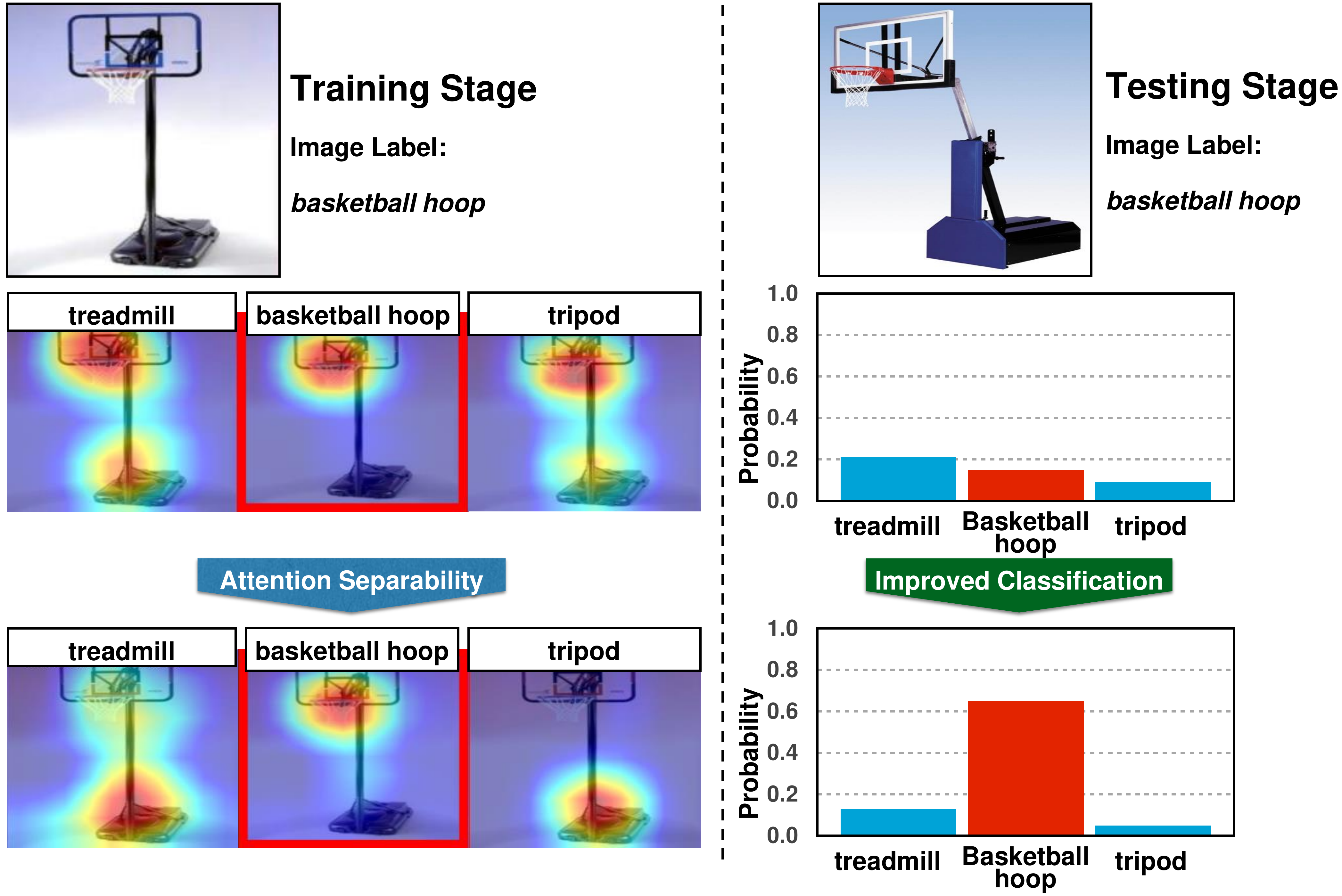}
    \caption{The baseline CNN attends to similar regions, \ie the central areas, when it comes to the relevant pixels for classes ``treadmill," ``basketball hoop," and ``tripod." The CNN with our proposed framework is able to tell the three classes apart and has high confidence to classify the input as ``basketball hoop."
    }
    \label{fig:main_idea}
\end{figure}

\begin{figure}[t]
    \centering
    \includegraphics[width=\columnwidth]{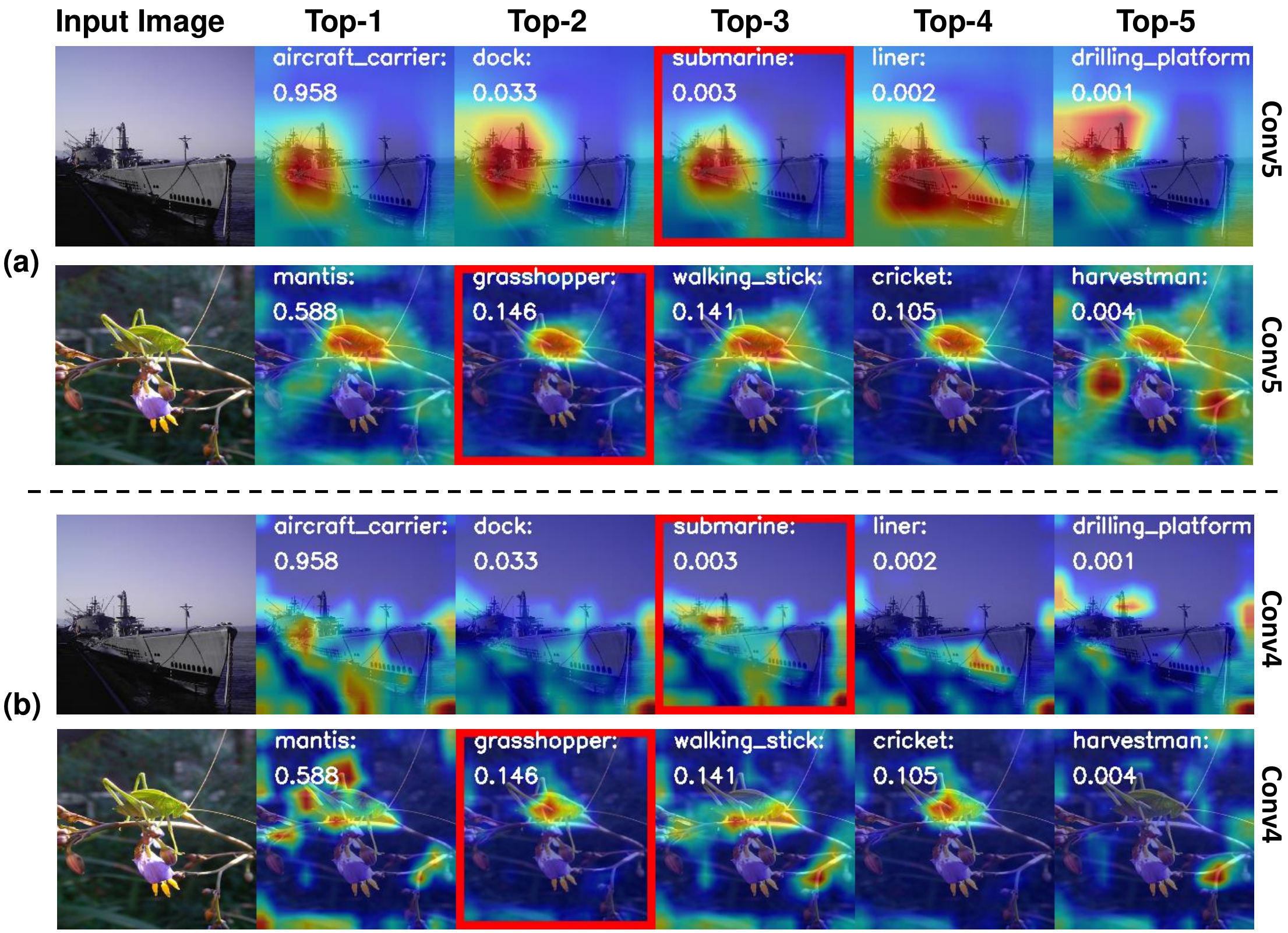}
    \caption{Grad-CAM~\cite{gradcam} attention maps of the VGG-19~\cite{simonyan2014very} top-5 predictions. Predictions with red-bounding boxes correspond to the ground-truth class. (a) Ground-truth class attention maps from the last layer (Conv5) have a large overlap with false positives (top-1 predictions). (b) Inner-layer attention maps (Conv4) are more separable than their last-layer counterparts.
    }
    \label{fig:att_overlap}
\end{figure}

Recently, class-specific attention has emerged as a powerful tool for interpreting CNNs \cite{zhou2016learning, gradcam, gradcamplus}. The big-picture intuition that drives these techniques is to answer the following question- \textsl{where is the target object in the image?} Some recent extensions \cite{li2018tell} make attention end-to-end trainable, producing attention maps with better localizability. While these methods consider the localization problem, this is insufficient for image classification, where the model needs to be able to tell various object classes apart. Specifically, existing methods produce attention maps corresponding to an individual class of interest that may not be \textsl{discriminative} across classes. Our intuition, shown in Figure~\ref{fig:main_idea}, is that such separable attention maps can lead to improved classification performance. Furthermore, we contend that false classifications stem from patterns across classes which confuse the model, and that eliminating these confusions can lead to better model discriminability. To illustrate this, consider Figure~\ref{fig:att_overlap} (a), where we use the VGG-19 model~\cite{simonyan2014very} to perform classification on the ILSVRC2012~\cite{ILSVRC15} dataset, we collect failure cases and generate the attention maps via Grad-CAM~\cite{gradcam} and we show the top-5 predictions. Figure~\ref{fig:att_overlap} (a) depicts that, while the attention maps of the last feature layer are reasonably well localized, there are large overlapping regions between the attention of the ground-truth class (marked by red bounding boxes) and the false positives, demonstrating the problem, and the need for \textsl{discriminative attention}.

To overcome the above attention-map limitations, we need to address two key questions: (a) \textsl{can we reduce visual confusion, i.e., make class-specific attention maps separable and discriminative across different classes?}, and (b) \textsl{can we incorporate attention discriminability in the learning process in an end-to-end fashion?} We answer these questions in a principled manner, proposing the first framework that makes attention maps class discriminative. Furthermore, we propose a new attention mechanism to guide model training towards attention discriminability, which provides end-to-end supervisory signals by explicitly enforcing attention maps of various classes to be separable.

Attention separability and localizability are key aspects of our proposed learning framework for image classification. Non-separable attention maps from the last layer, as shown in Figure~\ref{fig:att_overlap} (a), prompted us to look ``further inside" the CNN and Figure~\ref{fig:att_overlap} (b) shows attention maps from an intermediate layer. This illustration shows that these inner-layer attention maps are more separable than those from the last layer. However, the inner-layer attention maps are not as well-localized as the last layer. So, another question we ask is- \textsl{can we get the separability of the inner-layer attention and the localization of the last-layer attention at the same time?} Solving this problem would result in a ``best-of-both-worlds" attention map that is separable and localized, which is our goal. To this end, our framework also includes an explicit mechanism that enforces the ground-truth class attention to be cross-layer consistent.

We conduct extensive experiments on five competitive benchmarks (CIFAR-100~\cite{krizhevsky2009learning}, Caltech-256~\cite{griffin2007caltech}, ILSVRC2012~\cite{ILSVRC15}, CUB-200-2011~\cite{WahCUB_200_2011} and PASCAL VOC 2012~\cite{PASCAL-voc-2012}), showing performance improvements of 3.33\%, 1.64\%, 0.92\%, 4.8\%, and 5.73\%, respectively.

In summary, we make the following contributions:
\begin{itemize}
    \item We propose channel-weighted attention $\mathcal{A}_{ch}$, which has better localizability and avoids higher-order derivatives computation, compared to existing approaches for attention-driven learning.
    \item We propose attention separation loss $L_{AS}$, the first learning objective to enforce the model to produce class-discriminative attention maps, resulting in improved attention separability.
    \item We propose attention consistency loss $L_{AC}$, the first learning objective to enforce attention consistency across different layers, resulting in improved localization with ``inner-layer" attention maps.
    \item We propose ``Improving Classification with Attention Separation and Consistency" (ICASC), the first framework that integrates class-discriminative attention and cross-layer attention consistency in the conventional learning process. ICASC is flexible to be used with available attention mechanisms, \ie Grad-CAM~\cite{gradcam} and $\mathcal{A}_{ch}$, providing the learning objectives for training CNN with discriminative and consistent attention, which results in improved classification performance.
\end{itemize}

%------------------------------------------------------------------------
\section{Related work}
\textbf{Visualizing CNNs.}
Much recent effort has been expended in visualizing internal representations of CNNs to interpret the model. Erhan \etal~\cite{erhan2009visualizing} synthesized images to maximally activate a network unit. Mahendran \etal~\cite{mahendran2015understanding} and Dosovitskiy \etal~\cite{dosovitskiy2016inverting} analyzed the visual coding to invert latent representations, performing image reconstruction by feature inversion with an up-convolutional neural network. In ~\cite{simonyan2014deep, springenberg2015striving, zeiler2014visualizing}, the gradient of the prediction was computed \wrt the specific CNN unit to highlight important pixels. These approaches are compared in ~\cite{mahendran2016salient,gradcam}. The visualizations are fine-grained but not class-specific, where visualizations for different classes are nearly identical~\cite{gradcam}.

Our framework is inspired by recent works~\cite{zhou2016learning,gradcam,gradcamplus} addressing class-specific attention. CAM~\cite{zhou2016learning} generated class activation maps highlighting task-relevant regions by replacing fully-connected layers with convolution and global average pooling. Grad-CAM~\cite{gradcam} solved CAM's inflexibility where without changing the model architecture and retraining the parameters, class-wise attention maps were generated by means of gradients of the final prediction \wrt pixels in feature maps. However, we observe that directly averaging gradients in Grad-CAM~\cite{gradcam} results in the improper measurement of channel importance, producing substantial attention inconsistency among various feature layers. Grad-CAM++~\cite{gradcamplus} proposed to introduce higher-order derivatives to capture pixel importance, while its high computational cost in calculating the second- and third-order derivatives makes it impractical to be used during training.

\textbf{Attention-guided network training.}
Several recent methods~\cite{wang2017residualCVPR,hu2018squeeze,jetley2018learnICLR,woo2018cbam} have attempted to incorporate attention mechanisms to improve the performance of CNNs in image classification. Wang \etal~\cite{wang2017residualCVPR} proposed Residual Attention Network, modifying ResNet~\cite{he2016deep} by adding the hourglass network~\cite{newell2016stacked} to the skip-connection, generating attention masks to refine feature maps. Hu \etal~\cite{hu2018squeeze} introduced a Squeeze-and-Excitation (SE) module which used globally average-pooled features to compute channel-wise attention. CBAM~\cite{woo2018cbam,park2018bam} modified the SE module to exploit both spatial and channel-wise attention. Jetley \etal~\cite{jetley2018learnICLR} estimated attentions by considering the feature maps at various layers in the CNN, producing a 2D matrix of scores for each map. The ensemble of output scores was then used for class prediction. While these methods use attention for downstream classification, they do not explicitly use class-specific attention as part of model training for image classification.

\begin{figure}
    \centering
    %\vspace{-0.06em}
    \includegraphics[width=\columnwidth]{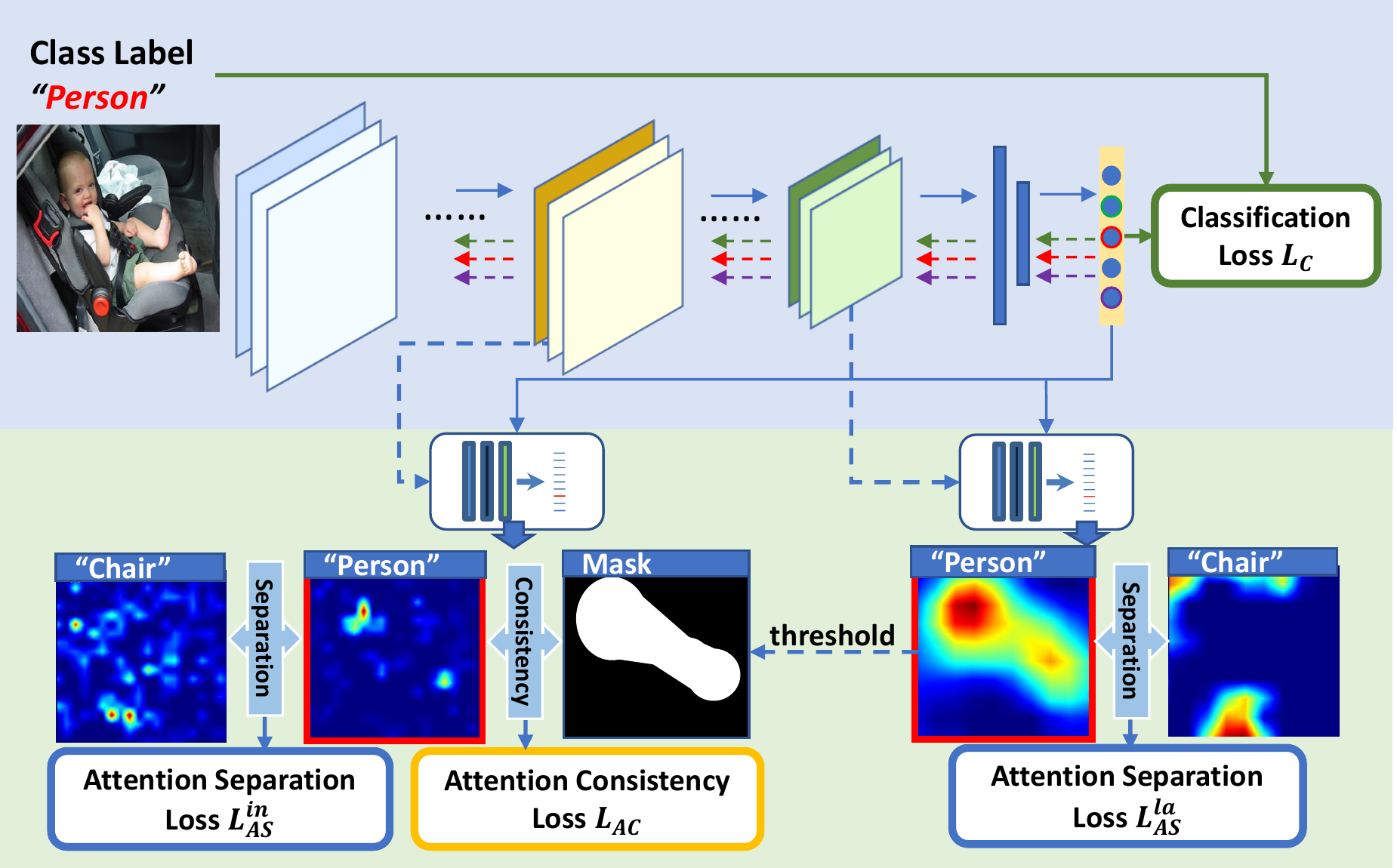}
    \caption{The framework of Improving Classification with Attention Separation and Consistency (ICASC).  }
    \label{fig:pipeline}
\end{figure}

Our work, to the best of our knowledge, is the first to use class-specific attention to produce supervisory signals for end-to-end model training with attention separability and cross-layer consistency. Furthermore, our proposed method can be considered as an add-on module to existing image classification architectures without needing any architectural change, unlike other methods~\cite{wang2017residualCVPR,hu2018squeeze,jetley2018learnICLR,woo2018cbam}. While class-specific attention has been used in the past for weakly-supervised object localization and semantic segmentation tasks~\cite{zhang2018adversarial,li2018tell,wei2017object,chaudhry2017discovering}, we model attention differently. The goal of these methods is singular - to make the attention well localize the ground-truth class, while our goal is two-fold - good attention localizability as well as discriminability. To this end, we devise novel objective functions to guide model training towards discriminative attention across different classes, leading to improved classification performance as we show in the experiments section.

\begin{figure}[]
    \centering
    \includegraphics[width=\columnwidth]{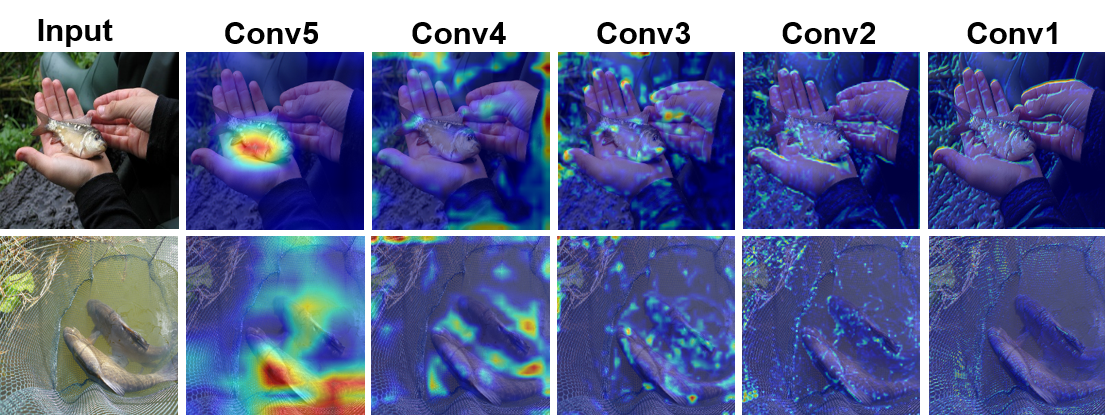}
    \caption{The Grad-CAM~\cite{gradcam} attentions of different VGG-19~\cite{simonyan2014very} feature layers for the 'tench' class. In both rows, the target is the fish while the model attention shifts across the layers.}
    \label{fig:badGradCam}
\end{figure}

\section{Approach}
In Figure~\ref{fig:pipeline}, we propose ``Improving Classification with Attention Separation and Consistency" (ICASC), the first end-to-end learning framework to improve model discriminability for image classification via attention-driven learning. The main idea is to produce separable attention across various classes, providing supervisory signals for the learning process. The motivation comes from our observations from Figure~\ref{fig:att_overlap} that the last layer attention maps computed by the existing methods such as Grad-CAM~\cite{gradcam} are not class-separable, although they are reasonably well localized. To address this problem, we propose the attention separation loss $L_{AS}$, a new attention-driven learning objective to enforce attention discriminability.

Additionally, we observe from Figure~\ref{fig:att_overlap} that inner layer attention at higher resolution has the potential to be separable, which suggests we consider both intermediate and the last layer attention to achieve separability and localizability at the same time. To this end, we propose the attention consistency loss $L_{AC}$, a new cross-layer attention consistency learning objective to enforce consistency among inner and last layer attention maps. Both proposed learning objectives require that we obtain reasonable attention maps from the inner layer. However, Grad-CAM~\cite{gradcam} fails to produce intuitively satisfying inner layer attention maps. To illustrate this, we depict two Grad-CAM~\cite{gradcam} examples in Figure~\ref{fig:badGradCam}, where we see the need for better inner layer attention. To this end, we propose a new channel-weighted attention mechanism $\mathcal{A}_{ch}$ to generate improved attention maps (explained in Sec.~\ref{subsec:A_ch}). We then discuss how we use them to produce supervisory signals for enforcing attention separability and cross-layer consistency.

%------------------------------------------------------------------------

\subsection{Channel-weighted attention $\mathcal{A}_{ch}$}
\label{subsec:A_ch}
Commonly-used techniques to compute gradient-based attention maps given class labels include CAM~\cite{zhou2016learning}, Grad-CAM~\cite{gradcam}, and Grad-CAM++~\cite{gradcamplus}. We do not use CAM because (a) it is inflexible, requiring network architecture modification and model re-training, and (b) it works only for the last feature layer.

Compared to CAM~\cite{zhou2016learning}, Grad-CAM~\cite{gradcam} and Grad-CAM++~\cite{gradcamplus} are both flexible in the sense that they only need to compute the gradient of the class prediction score \wrt the feature maps to measure pixel importance. Specifically, given the class score $Y^c$ for the class $c$ and the feature map $F^k$ in the $k$-th channel, the class-specific gradient is determined by computing the partial derivative $(\partial Y^c)/(\partial F^k )$. The attention map is then generated as $\mathcal{A}=ReLU(\sum_k\alpha_k^cF^k)$,
where $\alpha_k^c$ indicates the importance of $F^k$ in the $k$-th channel. In Grad-CAM~\cite{gradcam}, the weight $\alpha_k^c$ is a global average of the pixel importance in $(\partial Y^c)/(\partial F^k )$:
\vspace{-.5em}
\begin{equation}
    \alpha_k^c = \frac{1}{Z}\sum_i\sum_j\frac{\partial Y^c}{\partial F^k_{ij}}
    \label{eq:gradcamW}
\vspace{-.5em}
\end{equation}
where $Z$ is the number of pixels in $F^k$. Grad-CAM++~\cite{gradcamplus} further introduces higher-order derivatives to compute $\alpha_k^{c}$ so as to model pixel importance.

Although Grad-CAM~\cite{gradcam} and Grad-CAM++~\cite{gradcamplus} are more flexible than CAM~\cite{zhou2016learning}, they have several drawbacks that hinder their use \textsl{as is} for our purposes of providing separable and consistent attention guidance for image classification. First, there are large attention shifts among attention maps of different feature layers in Grad-CAM~\cite{gradcam} which are caused by negative gradients while computing channel-wise importance. A key aspect of our proposed framework ICASC is to exploit the separability we observe in inner layer attention in addition to good localization from the last layer attention. While we observe relatively less attention shift with Grad-CAM++~\cite{gradcamplus}, the high computational cost of computing higher-order derivatives precludes its use in ICASC since we use attention maps from multiple layers to guide model training in every iteration.

To address these issues, we propose channel-weighted attention $\mathcal{A}_{ch}$, highlighting the pixels where the gradients are positive. In our exploratory experiments, we observed that the cross-layer inconsistency of Grad-CAM~\cite{gradcam}, noted above, is due to negative gradients from background pixels. In Grad-CAM~\cite{gradcam}, all pixels of the gradient map contribute equally to the channel weight (Eq.~\ref{eq:gradcamW}). Therefore, in cases where background gradients dominate, the model tends to attend only to small regions of target objects, ignoring regions that are important for class discrimination.

We are motivated by prior work~\cite{gradcamplus,zeiler2014visualizing,springenberg2015striving} that observes that positive gradients \wrt each pixel in the feature map $F^k$ strongly correlate with the importance for a certain class. A positive gradient at a specific location implies increasing the pixel intensity in $F^k$ will have a positive impact on the prediction score, $Y^c$. To this end, driven by positive gradients, we propose a new channel-weighted attention mechanism $\mathcal{A}_{ch}$:

\vspace{-1.5em}
\begin{equation}
     \mathcal{A}_{ch} = \frac{1}{Z} ReLU(\sum_k  \sum_i\sum_j ReLU(\frac{\partial Y^c}{\partial F^k_{ij}}) F^k)
\label{eq:A_ch}
\vspace{-.5em}
\end{equation}

\begin{figure}[t]
    \centering
    \includegraphics[width=\columnwidth]{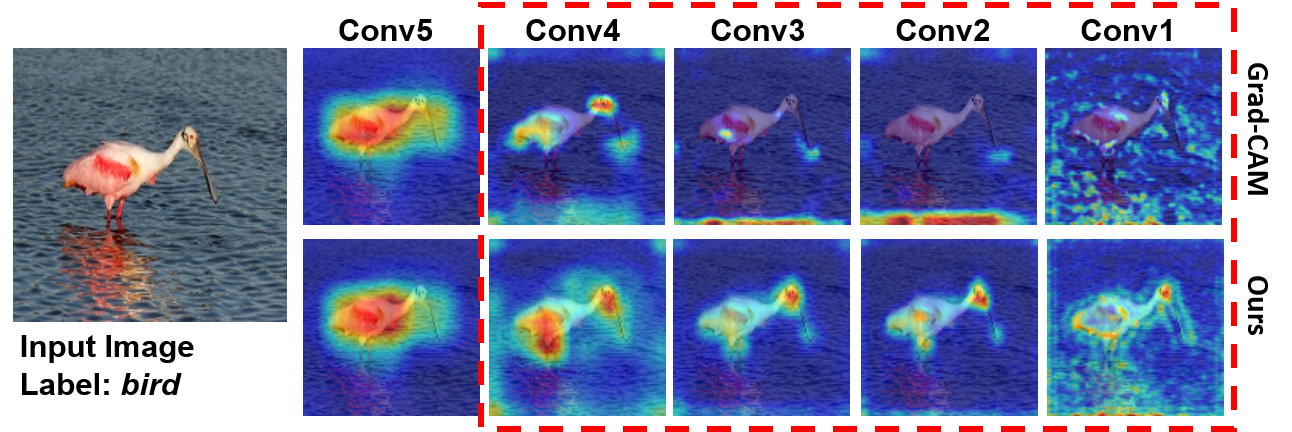}
    \caption{The comparison of attention maps from different VGG-19~\cite{simonyan2014very} layers. Ours has less attention shift than Grad-CAM~\cite{gradcam}. In the marked areas, ours attends to the target objects, \ie bird, while Grad-CAM~\cite{gradcam} tends to highlight the background pixels.}
    \label{fig:att_types}
\end{figure}

Our attention does not need to compute higher-order derivatives as in Grad-CAM++~\cite{gradcamplus}, while also resulting in well-localized attention maps with relatively less shift unlike Grad-CAM~\cite{gradcam}, as shown in Figure~\ref{fig:att_types}.

\subsection{Attention separation loss $L_{AS}$}
We use the notion of attention separability as a principled part of our learning process and propose a new learning objective $L_{AS}$. Essentially, given the attention map of a ground-truth class $\mathcal{A}^T$ and the most confusing class $\mathcal{A}^{Conf}$, where $\mathcal{A}^{Conf}$ comes from the non-ground truth class with the highest classification probability, we enforce the two attentions to be separable. We reflect this during training by quantifying overlapping regions between $\mathcal{A}^T$ and $\mathcal{A}^{Conf}$, and minimizing it. To this end, we propose $L_{AS}$ which is defined as:
\vspace{-.4em}
\begin{equation}
    L_{AS} = 2\cdot\frac{\sum_{ij}(min(\mathcal{A}^T_{ij}, \mathcal{A}^{Conf}_{ij})\cdot Mask_{ij})}{\sum_{ij}(\mathcal{A}^T_{ij} + \mathcal{A}^{Conf}_{ij})},
    \label{eq:SIoU}
\vspace{-.4em}
\end{equation}
where the $\cdot$ operator indicates scalar product, and $\mathcal{A}_{ij}^T$ and $A_{ij}^{Conf}$ represent the $(i,j)^{th}$ pixel in attention maps $\mathcal{A}^T$ and $\mathcal{A}^{Conf}$ respectively. The proposed $L_{AS}$ is differentiable which can be used for model training.

Additionally, to reduce noise from background pixels, we apply a mask to focus on pixels within the target object region for the $L_{AS}$ computation. In Eq.~\ref{eq:SIoU}, $Mask$ indicates the target object region generated by thresholding the attention map $\mathcal{A}^{T}$ from the last layer:
% \vspace{-.5em}
% \begin{equation}
%     Mask_{ij} = \frac{1}{1+exp(-\omega(\mathcal{A}^{last}_{ij} - \sigma))},
% \label{eq:mask}
% \vspace{-.5em}
% \end{equation}
\vspace{-.5em}
\begin{equation}
    Mask_{ij} = \frac{1}{1+exp(-\omega(\mathcal{A}^{T}_{ij} - \sigma))},
\label{eq:mask}
\vspace{-.5em}
\end{equation}
where we empirically choose values of $\sigma$ and $\omega$ to be $0.55\times max(\mathcal{A}^{T}_{ij})$ and 100 respectively.

\begin{figure}
    \centering
    \includegraphics[width=0.9\columnwidth]{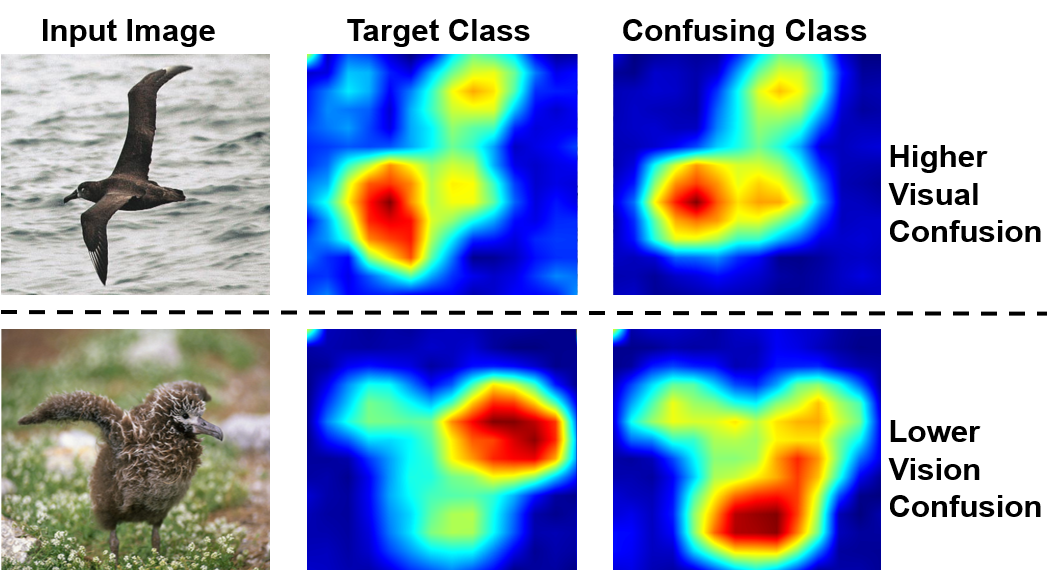}
    \caption{The top row demonstrates higher visual confusion than the bottom row. The two attention maps in the top row have high responses localized at the bird's head, while as shown in the bottom, the ground-truth class attention highlights the bird's head, and the confusing class attention addresses the lower body.}
    %\vspace{-0.5cm}
    \label{fig:sIoU}
\end{figure}

The intuition of $L_{AS}$ is illustrated in Figure~\ref{fig:sIoU}. If the model attends to the same or overlapped regions for different classes, it results in visual confusion. We penalize the confusion by explicitly reducing the overlap between the attention maps of the target and the most confusing class. Specifically, we minimize $L_{AS}$, which is differentiable with values ranging from 0 to 1.

The proposed $L_{AS}$ can be considered an add-on module for training a model without changing the network architecture. Besides applying $L_{AS}$ to the last feature layer, we can also compute $L_{AS}$ for any other layers, which makes it possible for us to analyze model attention at various scales.

While the proposed $L_{AS}$ helps enforce attention separability, it is not sufficient for image classification since inner layer attention maps are not as spatially well-localized as the last layer. We set out to achieve an attention map to be well-localized and class-discriminative, and to this end, we propose a new cross-layer attention consistency objective $L_{AC}$ that enforces the target attention map from an inner layer to be similar to that from the last layer.

\subsection{Attention consistency loss $L_{AC}$}
In higher layers (layers closer to output), the model attention captures more semantic information, covering most of the target object ~\cite{zhou2016learning,gradcam,gradcamplus}. For the intermediate layers with the smaller receptive fields of the convolution kernels, the model attends to more fine-grained patterns as shown in Figures~\ref{fig:badGradCam} and~\ref{fig:att_types}. Compared to higher-layer attention, lower-layer attention contains more noise, highlighting background pixels.

To address these issues, we propose the attention consistency loss $L_{AC}$ to correct the model attention so that the highlighted fine-grained attention is primarily localized in the target region:
\vspace{-.5em}
\begin{equation}
    L_{AC} = \theta -  \frac{\sum_{ij}(\mathcal{A}^{in}_{ij} \cdot Mask_{ij})}{\sum_{ij}\mathcal{A}^{in}_{ij}},
\vspace{-.5em}
\label{eq:lac}
\end{equation}
where $\mathcal{A}^{in}$ indicates attention maps from the inner feature layers, $Mask_{ij}$ (defined in Eq.~\ref{eq:mask}) represents the target region, and $\theta$ is set to 0.8 empirically. As can be noted from Eq.~\ref{eq:lac}, the intuition of $L_{AC}$ is that by exploiting last layer attention's good localizability, we can guide the inner layer attention to be chiefly concentrated within the target region as well. This guidance $L_{AC}$ helps maintain cross-layer attention consistency.

%------------------------------------------------------------------------
\subsection{Overall framework ICASC}
We apply the constraints of attention separability and cross-layer consistency jointly as supervisory signals to guide end-to-end model training, as shown in Figure~\ref{fig:pipeline}. Firstly, we compute inner-layer attention for the loss $L^{in}_{AS}$ with the purpose of enforcing inner-layer attention separability. For example, with ResNet, we use the last convolutional layer in the penultimate block. We empirically adopt this to compute $L^{in}_{AS}$ in consideration of the low-level patterns and semantic information addressed by the inner-layer attention. In Figure ~\ref{fig:att_types}, this inner-layer attention, with twice resolution as the last layer, highlights more fine-grained patterns while still preserving the semantic information, thus localizing the target object. We also apply the $L_{AS}$ constraint on the attention map from the last layer, giving us $L^{la}_{AS}$. Secondly, we apply the cross-layer consistency constraint $L_{AC}$ between the attention maps from these two layers. Finally, for the classification loss $L_C$, we use cross-entropy and multilabel-soft-margin loss for single and multi-label image classification respectively. The overall training objective of ICASC, $L$, is: %$L=L_C + L^{in}_{AS}+L^{la}_{AS}+L_{AC}$.
\vspace{-0.3em}
\begin{equation}
    L=L_C + L^{in}_{AS}+L^{la}_{AS}+L_{AC}
    \label{eq:overall}
\vspace{-0.3em}
\end{equation}
ICASC can be used with available attention mechanisms including Grad-CAM~\cite{gradcam} and $\mathcal{A}_{ch}$. We use ICASC$_{Grad-CAM}$ and ICASC$_{\mathcal{A}_{ch}}$ to refer to our framework used with Grad-CAM~\cite{gradcam} and $\mathcal{A}_{ch}$ as the attention mechanisms respectively.

%-------------------------------------------------------------------------
\section{Experiments}
Our experiments contain two parts, (a) evaluating the class discrimination of various attention mechanisms, and (b) demonstrating the effectiveness of the proposed ICASC by comparing it with the corresponding baseline model (having the same architecture) without the attention supervision. We conduct image classification experiments on various datasets, consisting of three parts: generic image classification on CIFAR-100 ($D_{CI}$)~\cite{krizhevsky2009learning}, Caltech-256 ($D_{Ca}$)~\cite{griffin2007caltech} and ILSVRC2012 ($D_I$)~\cite{ILSVRC15}, fine-grained image classification on CUB-200-2011 ($D_{CU}$)~\cite{WahCUB_200_2011}, and finally, multi-label image classification on PASCAL VOC 2012 ($D_P$)~\cite{PASCAL-voc-2012}. For simplicity, we use the shorthand in the parenthesis after the dataset names above to refer to each dataset and its associated task, and summarize all experimental parameters used in Table~\ref{tb:dataset_stats}. %, with additional details provided in the supplementary material.
We perform all experiments using PyTorch~\cite{paszke2017automatic} and NVIDIA Titan X GPUs. We use the same training parameters as those in the baselines proposed by the authors of the corresponding papers for fair comparison.

\begin{table}[t]
\centering
\small{
\begin{tabular}{@{}c@{}ccccc@{}}
\toprule
Task &$D_{CI}$ &$D_{Ca}$ &$D_I$ &$D_{CU}$ &$D_P$\\
\midrule
BNA &RN-18 &VGG &RN-18 &RN-50 &RN-18\\
 & &RN-18 & &RN-101 &\\
WD &5$e^{-4}$ &1$e^{-3}$ &1$e^{-4}$ &5$e^{-4}$ &1$e^{-3}$\\
MOM &0.9 &0.9 &0.9 &0.9 &0.9\\
LR &1$e^{-1}$ &1$e^{-2}$ &1$e^{-1}$ &1$e^{-3}$ &1$e^{-2}$\\
BS &128 &16 &256 &10 &16\\
OPM &SGD &CCA &SGD &SGD &CCA\\
\# epoch &160 &20 &90 &90 &20 \\
Exp. setting &\cite{he2016deep} &\cite{griffin2007caltech} &\cite{ILSVRC15} &\cite{sun2018multiECCV} &\cite{PASCAL-voc-2012}\\
\bottomrule
\end{tabular}
}
\vspace{.5em}
\caption{Experimental (exp.) settings used in this paper. VGG, RN-18, RN-50, and RN-101 denote VGG-19~\cite{simonyan2014deep}, ResNet-18~\cite{he2016deep}, ResNet-50, and ResNet-101, respectively. We use the same parameters as the references in the last row unless otherwise specified, putting more details in the appendix. Acronyms: BNA: base network architecture; WD: weight decay; MOM: momentum; LR: initial learning rate; BS: batch size; OPM: optimizer; SGD: stochastic gradient descent~\cite{bottou2010large}; CCA: cyclic cosine annealing~\cite{huang2017snapshot}.}
\vspace{-0.3cm}
\label{tb:dataset_stats}
\end{table}

\subsection{Evaluating class discriminability}
\begin{figure}[]
    \centering
    \includegraphics[width=0.9\columnwidth]{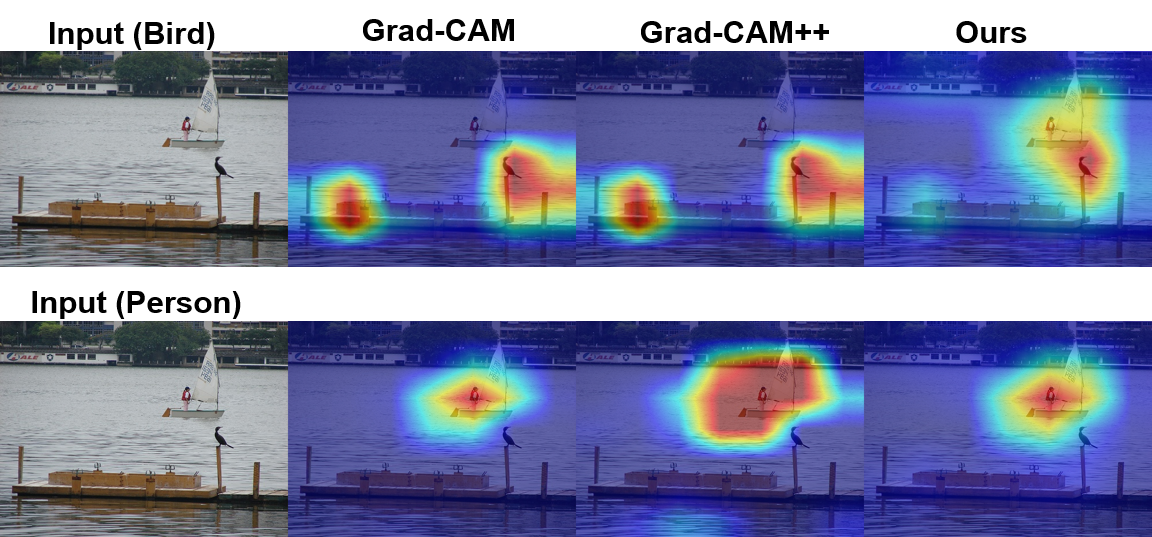}
    \caption{Multi-class attention maps (`bird' and `person').}
    \label{fig:att_pascal}
%\vspace{-.5em}
\end{figure}

\begin{figure}[]
    \centering
    \includegraphics[width=0.9\columnwidth]{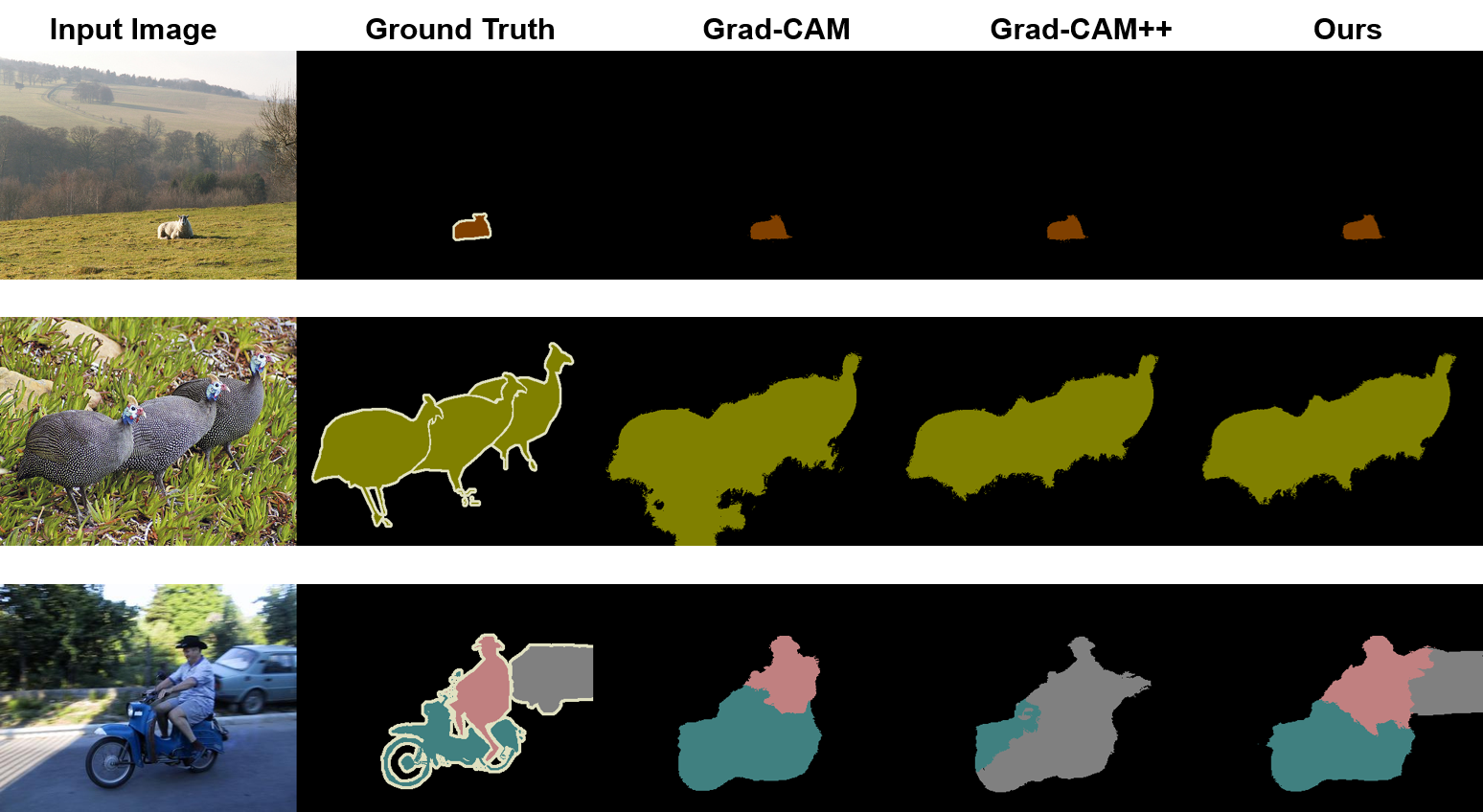}
    \caption{Segmentation masks generated from attention maps by DeepLab~\cite{deeplab} (best view in color, zoom in). From left to right: the Input Image, Ground Truth, Grad-CAM, Grad-CAM++ and ours.}
    \label{fig:seg}
\end{figure}

%\snote{Focus of this experiment is to show better class discriminativeness, and evaluation is not targeted specifically towards semantic segmentation. In Figure 8, show raw attention maps as well}
We first evaluate class-discriminability of our proposed attention mechanism $\mathcal{A}_{ch}$ by measuring both localizability (identifying target objects) and discriminability (separating different classes). We conduct experiments on the PASCAL VOC 2012 dataset. Specifically, with a VGG-19 model trained only with class labels (no pixel-level segmentation annotations), we generate three types of attention maps from the last feature layer: Grad-CAM, Grad-CAM++, and $\mathcal{A}_{ch}$. The attention maps are then used with DeepLab~\cite{deeplab} to generate segmentation maps, which are used to report both qualitative (Figure~\ref{fig:att_pascal} and~\ref{fig:seg}) and quantitative results (Table~\ref{tb:seg}), where we train Deeplab\footnote{https://github.com/tensorflow/models/tree/master/research/deeplab} in the same way as SEC~\cite{kolesnikov2016seed} is trained in~\cite{8733010}, using attention maps as weak localization cues. The focus of our evaluation here is targeted towards demonstrating class discriminability, and segmentation is merely used as a proxy task for this purpose.

Figure~\ref{fig:att_pascal} shows that $\mathcal{A}_{ch}$ (ours) has better localization for the two classes, ``Bird" and ``Person" compared to Grad-CAM and Grad-CAM++. In ``Bird," both Grad-CAM and Grad-CAM++ highlight false positive pixels in the bottom-left area, whereas in ``Person," Grad-CAM++ attends to a much larger region than Grad-CAM and $\mathcal{A}_{ch}$. Figure~\ref{fig:seg} qualitatively demonstrates better class-discriminative segmentation maps using $\mathcal{A}_{ch}$. In Figure~\ref{fig:seg} top row, as expected for a single object, all methods, including $\mathcal{A}_{ch}$, show good performance localizing the sheep. The second row shows that Grad-CAM covers more noise pixels of the grassland, while $\mathcal{A}_{ch}$ produces similar results as Grad-CAM++, both of which are better than Grad-CAM in identifying multiple instances of the same class. Finally, for multi-class images in the last row, $\mathcal{A}_{ch}$ demonstrates superior results when compared to both Grad-CAM and Grad-CAM++. Specifically, $\mathcal{A}_{ch}$ is able to tell the motorcycle, the person, and the car apart in the last row. We also obtain the quantitative results and report the score from the Pascal VOC Evaluation server in Table~\ref{tb:seg}, where $\mathcal{A}_{ch}$ outperforms both Grad-CAM and Grad-CAM++. The qualitative and quantitative results show that $\mathcal{A}_{ch}$ localizes and separates target objects better than the baselines, motivating us to use $\mathcal{A}_{ch}$ in ICASC, which we evaluate next.

\begin{table}
\centering
\small{
\begin{tabular}{@{\hspace{0mm}}l@{\hspace{2mm}}c@{\hspace{0mm}}}
\toprule
Attention Mechanism     &Score\\ \midrule
 Grad-CAM~\cite{gradcam} &56.65\\
 Grad-CAM++~\cite{gradcamplus} &51.70 \\
 %DeepLab~\cite{deeplab} + $\mathcal{A}_{px}$ & 57.69 \\
 $\mathcal{A}_{ch}$ (ours) &\bf{57.97} \\
 \bottomrule
\end{tabular}
}
\vspace{.5em}
\caption{Results on Pascal VOC 2012 segmentation validation set.}
%\vspace{-0.5cm}
%\znote{it is safer not to make the tables inline} %\snote{Modify table presentation. Mention that we use the attention maps generated by the attention mechanisms as the prior of the DeepLab~\cite{deeplab}.}}
\label{tb:seg}
\end{table}
%\knote{I'm not sure showing Gard-CAM++ result in Table~\ref{tb:seg} is helpful for our story. We implicitly assume that better attention maps lead to better segmentation results (as a proxy task), which is inconsistent with the first two rows of Table~\ref{tb:seg} (Grad-CAM vs. Grad-CAM++). As a reviewer, I'll automatically use the first two rows of Table~\ref{tb:seg} to argue that better mIoU doesn't mean better attention maps and question how meaningful Table~\ref{tb:seg} is given you mention that Grad-CAM++ produces better attention maps than Grad-CAM in Sec. 2.}
%\lnote{Here is what I quote from CVPR review ``Grad-CAM ++ modifies Grad-CAM by using different weights for different pixels, similar to the procedure mentioned in this paper. However this modification (in Gradcam++) leads to the visualization becoming not class-discriminative -- when multiple classes are present in the image the visualization fails to highlight just the class of interest. Hence it is important to show qualitative and quantitative results of the modified visualization technique on images with multiple classes." \\For \textbf{single type of objects}, Grad-CAM++ has the better mIoU than Grad-CAM (evaluted by Grad-CAM++ paper). The worse performance in table 2 is due to the multi-classes in VOC dataset. The purpose of the table 2 is to quantitatively validate that our modified attention works good both for single and multiple classes. And Fig.7 also qualitatively demonstrate it.}

\subsection{Evaluating $L_{AS}$ and $L_{AC}$ for image classification}
\subsubsection{Ablation study}
\begin{table}
\centering
%\begin{wraptable}{l}{4.8cm}
\small{
\begin{tabular}{l@{\hspace{2mm}}c@{\hspace{2mm}}c@{\hspace{2mm}}}
\toprule
 Method   &Top-1 &$\Delta$ \\ \midrule
 ResNet-50 & 81.70  &-\\
 + $L^{in}_{AS}$   &85.15  & 3.45\\
 + $L^{in}_{AS}$ + $L_{AC}$   &85.77  & 4.07\\
 %+ $L^{la}_{AS}$    & 85.78 & 4.08\\
 + $L^{in}_{AS}$ + $L^{la}_{AS}$ + $L_{AC}$    & \bf{86.20} &4.50\\
 \bottomrule
\end{tabular}
}
\vspace{0.5em}
\caption{Ablation study on CUB-200-2011 ($\Delta$=performance improvement; ``Top-1": top-1 accuracy (\%)).}
\vspace{-1em}
\label{tb:ablation}
\end{table}
%\end{wraptable}
Table~\ref{tb:ablation} shows an ablation study with the CUB-200-2011 dataset, which provides a challenging testing set given its fine-grained nature. We use the last convolutional layer in the penultimate block of ResNet-50 for computing $L^{in}_{AS}$ and the last layer attention map for $L^{la}_{AS}$. We see that $L^{in}_{AS}$ + $L^{la}_{AS} + L_{AC}$ achieves the best performance. The results show that the attention maps from the two different layers are complementary: last-layer attention has more semantic information, well localizing the target object, and inner layer attention with higher resolution provides fine-grained details. Though the inner-layer attention is more likely to be noisy than the last layer, $L_{AC}$ provides the constraint to guide the inner-layer attention to be consistent with that of the last layer and be concentrated within the target region. %\snote{Add a line or two discussing what we learn from this result.}

\begin{figure}[t]
    \centering
    \includegraphics[width=\columnwidth]{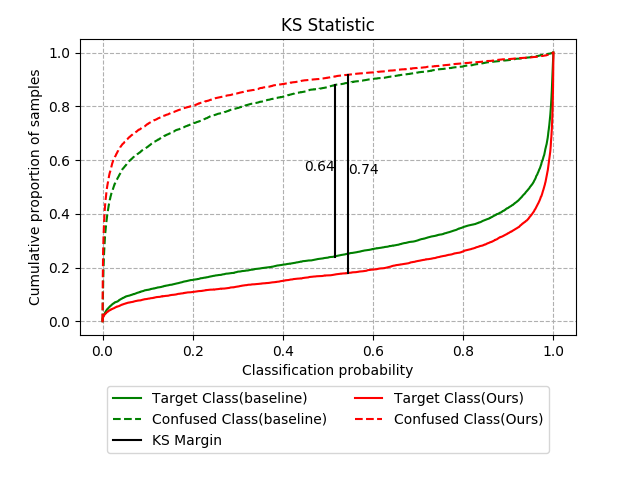}
    \vspace{-2.0em}
    \caption{The KS-Chart on the CUB-200-2011 testing set. ``Ours" stands for ResNet-50 + $L^{in}_{AS}$ + $L^{la}_{AS}$ + $L_{AC}$ in Table~\ref{tb:ablation}.} %More KS statistics can be found in the supplementary material.}
    %measures the degree of separation between the target and the most confusing classes distributions.
    \label{fig:ks_cub}
\end{figure}

%\snote{Add some refs that already use KS chart in vision, or at least emphasize that this is common for classification measurement.}
We quantitatively measure the degree of visual confusion reduction with our proposed learning framework. Specifically, as shown in Figure~\ref{fig:ks_cub}, we compute Kolmogorov-Smirnov (KS) statistics~\cite{ks} on the CUB-200-2011 testing set, measuring the degree of separation between the ground-truth (Target) class and the most confusing (Confused) class distributions~\cite{lopez2016revisiting}. We rank non-ground truth classes in descending order according to their classification probabilities and determine the most confusing class as the one ranked highest. In Figure~\ref{fig:ks_cub}, for the baseline model, the largest margin is 0.64 at the classification probability 0.51 whereas our proposed model has a KS margin of 0.74 at the classification probability 0.55. This demonstrates that our model is able to recognize $10\%$ more testing samples with higher confidence when compared to the baseline. %\snote{Report these numbers for at least one more dataset.}

%\begin{table}[]
%\centering
%\small{
%\begin{tabular}{@{\hspace{0mm}}l@{\hspace{1mm}}c@{\hspace{1mm}}c@{\hspace{0mm}}}
%\toprule
% Method   & Top-1 Acc. &$\Delta$ \\ \midrule
% ResNet-50 & 81.70  &-\\
% + $sIoU_{inner}$   &85.27  & 3.57\\
% + $sIoU_{last}$    & 85.78 & 4.08\\
% + $sIoU_{inner}$ + $sIoU_{last}$    & \bf{86.20} &4.50\\
% \bottomrule
%\end{tabular}
%}
%\vspace{.5em}
%\caption{The ablation study on the CUB-200-2011 dataset. $\mathcal{A}_{ch}$ is %used for $sIoU$ loss. The $Delta$ indicates top-1 accuracy improvement over the %vanilla ResNet-50. \snote{Experiment with and without Lac.} }
%\label{tb:ablation}
%\end{table}

\subsubsection{Generic image classification}

\begin{figure}[t]
    \centering
  \includegraphics[width=\columnwidth]{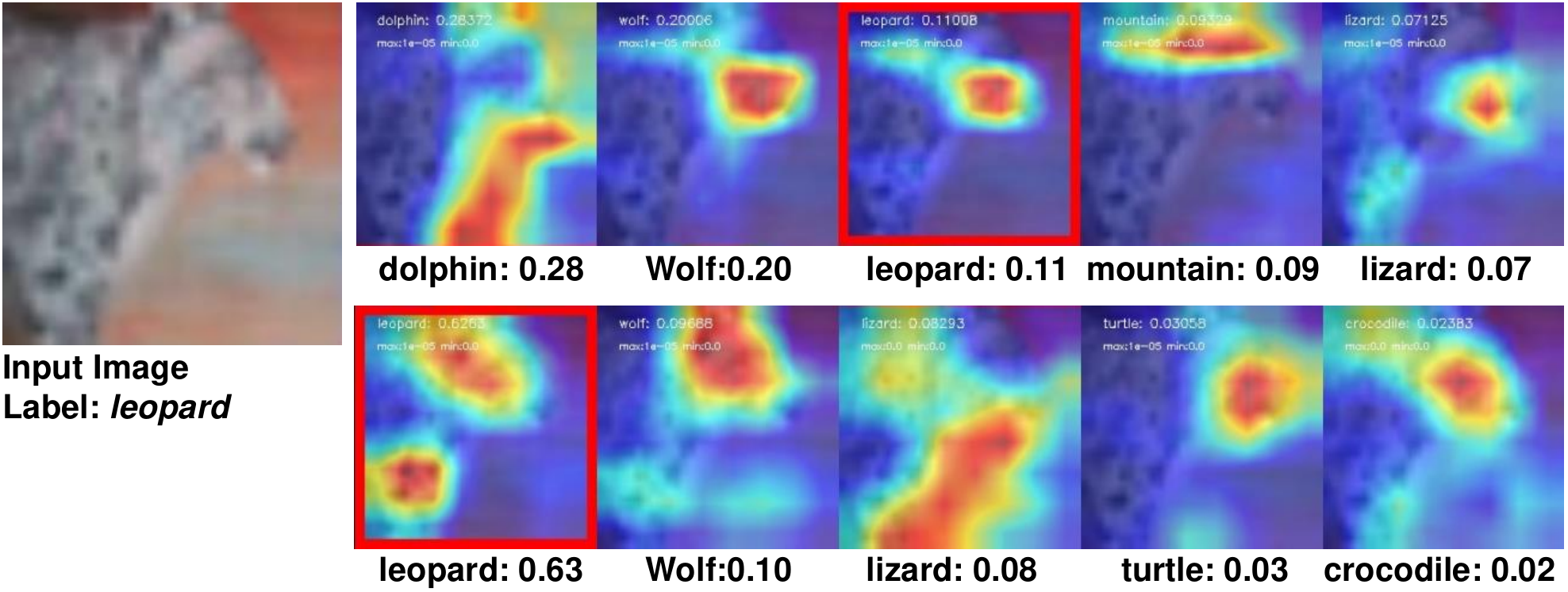}
    \caption{Qualitative results with CIFAR-100. We show top-5 predictions with classification scores given by ResNet-110 (top row) and ResNet-110 + ICASC$_{\mathcal{A}_{ch}}$ (bottom row).}
    \label{fig:cifar_qualitative}
\end{figure}

Tables~\ref{tb:cifar}-\ref{tb:imgnet} (in all tables, $\bigtriangleup$ indicates performance improvement of our method over baseline) show that the models trained with our proposed supervisory principles outperform the corresponding baseline models with a notable margin. %ICASC$_{Grad-CAM}$ means we use the same learning objective as Eq.~\ref{eq:overall}, but the mechanism to generate attention maps is Grad-CAM.
The most noticeable performance improvements are observed with the CIFAR-100 dataset in Table~\ref{tb:cifar}, which shows that, without changing the network architecture, the top-1 accuracy of ResNet-110 with our proposed supervision outperforms the baseline model by 3.33\%. Our supervised ResNet-110 also outperforms the one with stochastic depth and even the much deeper model with 164 layers. As can be observed from the qualitative results in Figure~\ref{fig:cifar_qualitative}, %(please see supplementary material for more results),
ICASC$_{\mathcal{A}_{ch}}$ equips the model with discriminative attention where the ground-truth class attention is separable from the confusing class, resulting in improved prediction.

%As for the comparison between pixel and channel wise attention, pixel-wise attention $\mathcal{A}_{px}$, which provides the sharper high-frequency attentive regions, is better in capturing discriminative parts in low resolution images when comapared to $\mathcal{A}_{ch}$.

\begin{table}[]
\centering
\small{
\begin{tabular}{@{\hspace{0mm}}l@{\hspace{2mm}}c@{\hspace{2mm}}c@{\hspace{0mm}}}
\toprule
 Method   &Top-1 &$\bigtriangleup$ \\ \midrule
 ResNet-110~\cite{huang2016deep}   & 72.78 & -\\
 ResNet-110 with Stochastic Depth~\cite{huang2016deep}   & 75.42 &- \\
 ResNet-164 (pre-activation)~\cite{huang2016deep}   & 75.63 & -
 \\\midrule
 ResNet-110 + ICASC$_{Grad-CAM}$    & 74.02 & 1.24 \\
 ResNet-110 + ICASC$_{\mathcal{A}_{ch}}$    & \bf{76.11} & \bf{3.33} \\
\bottomrule
\end{tabular}
}
\vspace{.5em}
\caption{Image classification results on CIFAR-100.}% \knote{Do we really need the ``Depth" column here? The depth is already indicated in the network name (ResNet-depth).}} %\knote{Why do we need the column ``Params" here (it is the same for all methods)? I assume all the ResNet here is ResNet-110, so I add ``-110" to the second and third methods for clarity. Please confirm whether that's the case. If yes, please remove my comment here.}}
\label{tb:cifar}
\end{table}

\begin{table}[]
\centering
\small{
%\begin{tabular}{@{\hspace{0mm}}c@{\hspace{2mm}}c@{\hspace{0mm}}}
\begin{tabular}{@{\hspace{0mm}}l@{\hspace{2mm}}c@{\hspace{2mm}}c@{\hspace{1mm}}c@{\hspace{5mm}}c@{\hspace{2mm}}c@{\hspace{1mm}}c@{}}
\toprule
\multirow{2}{*}{Method} & &N=30 & & &N=60 &\\
%\midrule
  &Top-1 &Top-5 &$\Delta$ &Top-1 &Top-5 &$\Delta$\\ \midrule[0.75pt]
 RN-18~\cite{he2016deep} &76.77 &92.48  & - &80.01 &94.12 &-\\
% %ResNet-18 + $\mathcal{A}_{px}$  &77.82  &92.43 &1.05\\
 RN-18 + ICASC$_{\mathcal{A}_{ch}}$  &\bf{78.01}  &\bf{92.87} &1.24 &\bf{81.32} &\bf{94.57} &1.31\\
\midrule
 VGG-19~\cite{simonyan2014very} &74.52 &90.05 & - &78.16 &92.17 & -\\
% %VGG-19 + $\mathcal{A}_{px}$ &75.34 &90.82 &0.82 \\
 VGG-19 + ICASC$_{\mathcal{A}_{ch}}$ &\bf{75.60} &\bf{90.85} &1.08 & \bf{79.80} &\bf{93.25} &1.64
 \\\bottomrule
 \end{tabular}
 %&
 %\midrule[1pt]
%\begin{tabular}{@{}lccc@{}}
%\toprule[1pt]
% N=60 &Top-1 &Top-5 &$\bigtriangleup$ \\ \midrule
% \cite{he2016deep} &80.01 &94.12 &-\\
% %ResNet-18 + $\mathcal{A}_{px}$  &\bf{81.34}  &94.20 &1.33\\
% \cite{he2016deep} + ICASC &81.32 &\bf{94.57} &1.31\\
%\midrule
% \cite{simonyan2014deep}  &78.16 &92.17 & -\\
% %VGG-19 + $\mathcal{A}_{px}$ &79.53 &93.22 &1.37 \\
% \cite{simonyan2014deep} + ICASC & \bf{79.80} %&\bf{93.25} &1.64 \\
%\bottomrule[1pt]
%\end{tabular}
%\end{tabular}
}
\vspace{.5em}
\caption{Results on Caltech-256. ``Top-5": top-5 accuracy (\%). ``RN-18": ResNet-18. ``N": \# of training images per class. We follow~\cite{griffin2007caltech} to randomly select 30 or 60 training images per class.}
\label{tb:caltech}
\end{table}

\begin{table}[]
\centering
\small{
\begin{tabular}{@{}lccc@{}}
\toprule[1pt]
Method &Top-1 &Top-5 &$\bigtriangleup$ \\ \midrule
 ResNet-18~\cite{he2016deep} &69.51 & 88.91 & -\\
 %ResNet-18 + ICASC_{Grad-CAM} &69.84 & - &0.33\\
 ResNet-18 + ICASC$_{\mathcal{A}_{ch}}$ &\bf{69.90} & \bf{89.71} &0.39\\
\midrule
 ResNet-18 + tenCrop~\cite{he2016deep} &72.12 & 90.58 & -\\
 ResNet-18 + tenCrop + ICASC$_{\mathcal{A}_{ch}}$ &\bf{73.04} & \bf{90.65} &0.92\\
\bottomrule[1pt]
\end{tabular}
}
\vspace{.5em}
\caption{Results on ILSVRC2012. %\lnote{\cite{he2016deep} reports the performance of ResNet18 and Resnet18+tencrop. I put \cite{he2016deep} here to indicate the accuracy of Resnet18+tencrop is from that paper.} \knote{Got it. I've already changed the table accordingly.}
}
%\vspace{-2em}
\label{tb:imgnet}
\end{table}

\subsubsection{Fine-grained Image Recognition}
For fine-grained image recognition, we evaluate our approach on the CUB-200-2011 dataset~\cite{WahCUB_200_2011}, which contains 11788 images (5994/5794 for training/testing) of 200 bird species. We show the results in Table~\ref{tb:cub}. We observe that training with our learning mechanism boosts the accuracy of the baseline ResNet-50 and ResNet-101 by 4.8\% and 4.0\% respectively. Our method achieves the best overall performance against the state-of-the-art. Furthermore, with ResNet-50, our method outperforms even the method that uses extra annotations (PN-CNN) by 0.8\%. %The fact that our method outperforms all of the methods that use extra annotation demonstrates that with our principled attention guidance, the model attends are able to look at the discriminative regions of the target objects automatically without necessity of high-cost labelling.

ICASC$_{\mathcal{A}_{ch}}$ has better flexibility compared to the other methods in Table~\ref{tb:cub}. The existing methods are specifically designed for fine-grained image recognition where, according to prior knowledge of the fine-grained species, the base network architectures (BNA) are modified to extract features of different objects parts~\cite{zhang2016spda,zheng2017learning,sun2018multiECCV}. In contrast, ICASC$_{\mathcal{A}_{ch}}$ needs no prior knowledge and works for generic image classification without changing the BNA.

\begin{table}[]
\centering
\small{
\begin{tabular}{@{}l@{}c@{\hspace{2mm}}c@{\hspace{2mm}}c@{\hspace{2mm}}c@{}}
\toprule
 Method     &  No Extra Anno.     &  1-Stage  &Top-1 &$\Delta$\\ \midrule
 ResNet-50~\cite{sun2018multiECCV}  & \green{\ding{51}} & \green{\ding{51}} & 81.7 &-\\
 ResNet-101~\cite{sun2018multiECCV} & \green{\ding{51}} & \green{\ding{51}} & 82.5 &0.8\\
 MG-CNN~\cite{wang2015multiple} &  \red{\ding{56}} & \red{\ding{56}} & 83.0 &1.3\\
 SPDA-CNN~\cite{zhang2016spda}   & \red{\ding{56}} & \green{\ding{51}} & 85.1 &3.4\\
 RACNN~\cite{fu2017look}      &  \green{\ding{51}} & \green{\ding{51}} & 85.3 &3.6\\
 PN-CNN~\cite{branson2014birdBMVC}     & \red{\ding{56}} & \red{\ding{56}} & 85.4 &3.7\\
 RAM~\cite{li2017dynamicICCV}        & \green{\ding{51}} & \red{\ding{56}} & 86.0 &4.3\\
 MACNN + 2parts~\cite{zheng2017learning} & \green{\ding{51}} & \green{\ding{51}} & 85.4 &3.7\\
 %MACNN + 4parts~\cite{zheng2017learning} &  \ding{56} & \ding{51} & \bf{86.5} \\
 ResNet-50 + MAMC~\cite{sun2018multiECCV}  & \green{\ding{51}} & \green{\ding{51}} & 86.2 &4.5\\
 ResNet-101 + MAMC~\cite{sun2018multiECCV}  & \green{\ding{51}} & \green{\ding{51}} & \bf{86.5} &4.8
 \\\midrule
 ResNet-50 + ICASC$_{\mathcal{A}_{ch}}$  & \green{\ding{51}} & \green{\ding{51}} &  86.2 &4.5\\ %86.16\\
 ResNet-101 + ICASC$_{\mathcal{A}_{ch}}$  & \green{\ding{51}} & \green{\ding{51}} & \bf{86.5} &4.8\\%86.47\\
\bottomrule
\end{tabular}
}
\vspace{.5em}
\caption{Results on CUB-200-2011. ``No Extra Anno." means not using extra annotation (bounding box or part) in training. ``1-Stage" means the training is done in one stage.}
%\vspace{-0.5cm}
\label{tb:cub}
\end{table}

\subsubsection{Multi-class Image Classification}
We conduct multi-class image classification on the PASCAL VOC 2012 dataset, which contains 20 classes. Different from the above generic and fine-grained image classification where each image is associated with one class label, for each of the 20 classes, the model predicts the probability of the presence of an instance of that class in the test image. As our attention is class-specific, we can seamlessly adapt our pipeline from single-label to multi-label classification. Specifically, we apply the one-hot encoding to corresponding dimensions in the predicted score vector and compute gradients to generate the attention for multiple classes. As for the most confusing class, we consistently determine it as the non-ground truth class with the highest classification probability.

For evaluation, we report the Average Precision (AP) from the PASCAL Evaluation Server~\cite{PASCAL-voc-2012}. We also compute the AUC score via scikit-learn python module~\cite{scikit-learn} as an additional evaluation metric~\cite{binder2012taxonomies}. Table~\ref{tb:voc} shows that ResNet-18~\cite{he2016deep} with $\mathcal{A}_{ch}$ outperforms the baseline by 5.73\%. %Furthermore, $\mathcal{A}_{ch}$ also helps ResNet-18~\cite{he2016deep} outperform the deeper model ResNet-34~\cite{leaderboard} in AP.

\begin{table}[]
\centering
\small{
\begin{tabular}{@{}lccc@{}}
\toprule
 Method  &AUC Score & AP (\%)    &$\bigtriangleup$ \\ \midrule
 ResNet-18~\cite{he2016deep}                       & 0.976  &  77.44 & -\\
 %ResNet-18 + ICASC_{Grad-CAM}  & -  & 82.12 &  4.68\\
 ResNet-18 + ICASC$_{\mathcal{A}_{ch}}$  & \bf{0.981}  & \bf{83.17} &  5.73\\
% \midrule
% ResNet-34~\cite{leaderboard}     & -      & 80.70 & -\\
% ResNet-34~\cite{he2016deep} + ICASC & \bf{0.983} & \bf{86.49}  & 5.79\\
\bottomrule
\end{tabular}
}
\vspace{.5em}
\caption{Results on Pascal VOC 2012. %The top-1 accuracy pf ResNet-34 is cited from the PASCAL leader-board.
}
\label{tb:voc}
\end{table}

%\subsubsection{Comparing attention mechanisms for supervising image classification}
\subsubsection{Comparing attention mechanisms}
%In this section, we compare the image classification performance where the models are trained with supervision from different attention mechanisms, including Grad-CAM, $\mathcal{A}_{ch}$ and $\mathcal{A}_{px}$. According to Table~\ref{tb:cifar_gradcam} and~\ref{tb:cls}, the higher accuracy achieved by $\mathcal{A}_{ch}$ and $\mathcal{A}_{px}$ demonstrates that our proposed attention mechanisms give the better supervision for model training than Grad-CAM. Additionally, we note that even the model trained with Grad-CAM as the guiding mechanism outperforms the baseline, further validating our contribution of attention-driven learning for reduced visual confusion.

We compare the image classification performance when ICASC is trained with Grad-CAM~\cite{gradcam} and $\mathcal{A}_{ch}$. As can be noted from the results in Table~\ref{tb:cifar} and~\ref{tb:cls}, the higher Top-1 accuracy of ICASC$_{\mathcal{A}_{ch}}$ shows that our attention mechanism provides better supervisory signals for model training than Grad-CAM~\cite{gradcam}. Additionally, even ICASC with Grad-CAM still outperforms the baseline, further validating our key contribution of attention-driven learning for reducing visual confusion. The proposed ICASC is flexible to be used with any existing attention mechanisms as well, while resulting in improved classification performance.

\begin{table}[]
\centering
\small{
\begin{tabular}{@{\hspace{0mm}}c@{\hspace{2mm}}c@{\hspace{0mm}}}
\begin{tabular}{@{\hspace{0mm}}l@{\hspace{2.5mm}}c@{\hspace{2mm}}}
\toprule
 Pascal VOC 2012   &Top-1  \\ \midrule
 ResNet-18 & 77.44  \\
 + ICASC$_{Grad-CAM}$   &82.12  \\
 %+ $\mathcal{A}_{px}$    & \bf{83.22} & 5.78\\
 + ICASC$_{\mathcal{A}_{ch}}$    & \bf{83.17}\\
\bottomrule[0.5pt]
\end{tabular}
&
\begin{tabular}{@{\hspace{0mm}}l@{\hspace{2mm}}c@{\hspace{2mm}}}
\toprule
 Caltech-256 &Top-1   \\ \midrule
 ResNet-18 & 80.01 \\
 + ICASC$_{Grad-CAM}$   &80.28   \\
 %+ $\mathcal{A}_{px}$   & \bf{81.34} &1.33\\
 + ICASC$_{\mathcal{A}_{ch}}$   & \bf{81.32}\\
 %VGG-19 + ICASC_{Grad-CAM}   &78.61  &-  \\
 %VGG-19 + $\mathcal{A}_{px}$   & 79.53 &- \\
 %VGG-19 + ICASC   & 79.80 &- \\
\bottomrule[0.5pt]
\end{tabular}
\\
\begin{tabular}{@{\hspace{1mm}}l@{\hspace{2mm}}c@{\hspace{2mm}}}
\toprule[0.5pt]
 CUB-200-2011 &Top-1   \\ \midrule
 ResNet-50 & 81.70  \\
 + ICASC$_{Grad-CAM}$   &85.45    \\
 %ResNet-50 + $\mathcal{A}_{px}$   & \red{TBD} & - \\
 + ICASC$_{\mathcal{A}_{ch}}$   & \bf{86.20}  \\
\bottomrule
\end{tabular}
&
\begin{tabular}{@{\hspace{0mm}}l@{\hspace{2mm}}c@{\hspace{2mm}}}
 \toprule[0.5pt]
 ILSVRC2012  &Top-1   \\ \midrule
 ResNet-18 & 69.51\\
 + ICASC$_{Grad-CAM}$   & 69.84    \\
 %ResNet-18 + $\mathcal{A}_{px}$   & \red{TBD} & - \\
 + ICASC$_{\mathcal{A}_{ch}}$   & \bf{69.90} \\
 %ResNet-18 + ICASC_{Grad-CAM} + tenCrop  & \red{TBD}  &-  \\
 %ResNet-18 + $\mathcal{A}_{px}$ +tenCrop   & \red{TBD} & - \\
 %ResNet-18 + ICASC +tenCrop   & 73.04 & - \\
 \bottomrule
\end{tabular}
\end{tabular}
}
\vspace{.5em}
\caption{Comparing baseline, ICASC$_{Grad-CAM}$ and ICASC$_{\mathcal{A}_{ch}}$. %``Acc." means Top-1 accuracy. %The column $\Delta$ indicates the performance improvement over the Grad-CAM
}
\vspace{-0.1cm}
\label{tb:cls}
\end{table}

\section{Conclusions}
We propose a new framework, ICASC, which makes class-discriminative attention a principled part of training a CNN for image classification. Our proposed attention separation loss and attention consistency loss provide supervisory signals during training, resulting in improved model discriminability and reduced visual confusion. Additionally, our proposed channel-weighted attention has better class discriminability and cross-layer consistency than existing methods (e.g. Grad-CAM~\cite{gradcam}). ICASC is applicable to any trainable network without changing the architecture, giving an end-to-end solution to reduce visual confusion. ICASC achieves performance improvements on various medium-scale, large-scale, fine-grained, and multi-class classification tasks. While we select last two feature layers which contain most semantic information to generate the attention maps, ICASC is flexible \wrt layer choices for attention generation, and we plan to study the impact of various layer choices in the future.

%We proposed a trainable class discriminative attention mechanism to guide model training for image classification. Our proposed framework has better attention consistency among different layers than Grad-CAM~\cite{gradcam},
%and it is flexible to be added to any trainable network without changing network architectures, giving an end-to-end procedure to reduce model visual confusion. We show performance improvements on various medium-scale, large-scale, fine-grained and multi-class classification tasks.

%We proposed a trainable class-wise attention to guide model training for image classification. The proposed attention mechanism has less computation cost than Grad-CAM++~\cite{gradcamplus} and better attention consistency among different layers than Grad-CAM~\cite{gradcam}. The proposed attention guided training is flexible to be added to any networks without changing net architectures, which is an end-to-end procedure, reducing the visual confusion.  The conducted experiments shows  performance boosting on various image-classification tasks, including classifying images in the setting of medium scale, large-scale, fine-grained and multi-class.
%-------------------------------------------------------------------------

%{\small
%\bibliographystyle{ieee_fullname}
%\bibliography{egbib}
%}

%\end{document}

%%%%%%%%%%%%%%%%%%%%%%%%%KC
%%%%%%%%%%%%%%%%%%%%%%%%%KC
%%%%%%%%%%%%%%%%%%%%%%%%%KC

\clearpage

%\begin{document}

%%%%%%%%% TITLE
\title{Sharpen Focus: Learning with Attention Separability and Consistency \\Supplementary Material}
%\title{Lean a Better Visual Explanation: Attention-Guided Training for Image Classification with Visual Confusion Reducing}

% \and

% \author{Lezi Wang$^{1}$, Ziyan Wu$^{2}$, Srikrishna Karanam$^{2}$, Kuan-Chuan Peng$^{2}$,\\ Rajat Vikram Singh$^{2}$, Bo Liu$^{1}$, and Dimitris N. Metaxas$^{1}$\\
% $^{1}$Rutgers University, New Brunswick NJ\\
% $^{2}$Siemens Corporate Technology, Princeton NJ\\
% { \small \{lw462, lb507, dnm\}@cs.rutgers.edu,}
% { \small \{ziyan.wu, srikrishna.karanam, kuanchuan.peng, singh.rajat\}@siemens.com}
% }

%\maketitle
%\thispagestyle{empty}

%\knote{Line numbers in the main paper where we refer to the supplementary material: (1) L529, L554: more details about the experimental settings. (2) L665: more KS statistics. (3) L744: more qualitative results (similar to Figure 10) and discussion.}

%KC: reset the section counter
\setcounter{section}{0}

\begin{appendices}
\section{Implementation details}
\begin{table*}[t]
\renewcommand\thetable{1}
\centering
\small{
\begin{tabular}{c@{\hspace{3mm}}c@{\hspace{3mm}}c@{\hspace{3mm}}c@{\hspace{3mm}}c@{\hspace{3mm}}c}
\toprule
dataset &CIFAR-100~\cite{krizhevsky2009learning} &Caltech-256~\cite{griffin2007caltech} &ILSVRC2012~\cite{ILSVRC15} &CUB-200-2011~\cite{WahCUB_200_2011} &PASCAL VOC 2012~\cite{PASCAL-voc-2012}\\
\midrule
\# classes &100 &256 &1000 &200 &20\\
image size &32$\times$32 &299$\times$299 &224$\times$224 &448$\times$448 &299$\times$299\\
\# images &60000 &30607 &$\sim$1.3M &11788 &15000\\
\midrule
\# training images &50000 &7680/15360 &1.2M &5994 &5717\\
\# testing images &10000 &6400 &50000 &5794 &5823\\
%data split &specified \red{\cite{}} &random %&specified \red{\cite{}} &specified %\red{\cite{}} &specified \red{\cite{}}\\
training batch size &128 &16 &256 &10 &16\\
weight decay &$0.0005$ &$10^{-3}$ &$10^{-4}$ &$10^{-4}$ &$10^{-3}$\\
momentum &0.9 &0.9 &0.9 &0.9 &0.9\\
initial learning rate &0.1 &0.01 &0.1 &$10^{-3}$ &0.01\\
%lr adjustment &step_decay &0.01 &0.1 %&$10^{-3}$ &0.01\\
\# training epochs &160 &20 &90  &90 &20\\
evaluation metric &Top-1 Accuracy &Top-1 Accuracy &Top-1 Accuracy &Top-1 Accuracy &mean Average Precision\\
%reference of above setting &~\cite{he2016deep} & ~\cite{griffin2007caltech} &~\cite{huang2017densely} &\cite{liu2016fully,sun2018multiECCV} &~\cite{leaderboard,PASCAL-voc-2012}\\
\bottomrule
\end{tabular}
\vspace{.5em}
\caption{The details of the dateset and training parameters.}
%\caption{The details of the dateset and training parameters. The references of training recipes are listed at the bottom.}
\label{table:dataset}
}
\end{table*}
%In this section, we first conduct the ablation study to investigate the impact of each component on the performance. Then we evaluate our approach on five benchmarks with the optimal settings from the ablation analysis.
As noted in the main paper, our experiments contain three parts: (a) generic image classification evaluation on three datasets:  CIFAR-100~\cite{krizhevsky2009learning},  Caltech-256~\cite{griffin2007caltech} and ILSVRC2012~\cite{ILSVRC15}, (b) fine-grained image classification evaluation on
the CUB-200-2011~\cite{WahCUB_200_2011} dataset, and (c) multi-label image classification evaluation on the PASCAL VOC 2012~\cite{PASCAL-voc-2012} dataset. We perform experiments using PyTorch~\cite{paszke2017automatic} and
NVIDIA Titan X GPUs. We do not search in the hyperparameter space for the best hyperparameters and instead use the same training parameters as those in the corresponding baselines. Complete experimental details about training and the five datasets we used are provided in Table~\ref{table:dataset}.

\subsection{Generic image classification}
\textbf{CIFAR-100}: The image is padded by 4 pixels on each side, filled with 0 value resulting in a 40$\times$40 image. A 32$\times$32 crop is randomly sampled from an image or its horizontal flip, with the per-pixel RGB mean value subtracted. We adopt the same weight initialization method following ~\cite{he2016deep} and train the ResNet using Stochastic Gradient Descent (SGD)~\cite{bottou2010large} with a mini-batch size of 128. We use a weight decay of 0.0005 with a momentum of 0.9 and set the initial learning rate to 0.1. The learning rate is divided by 10 at 81 and 122 epochs. The training is terminated after 160 epochs.

\textbf{Caltech-256}: There is no official training/testing data split. We follow the work in ~\cite{griffin2007caltech} to randomly select 25 images per category as the testing set and 30, 60 images per category as training. We remove the last (257-th) category ``clutter," keeping the 256 categories which describe specific objects. We use VGG-19~\cite{simonyan2014deep} and ResNet-18~\cite{he2016deep} as the baseline models. For the training of both the baseline and our proposed method, we use a weight decay of 0.001 with a momentum of 0.9 and set the initial learning rate to 0.01. To speed up the model training, we adopt cyclic cosine annealing~\cite{huang2017snapshot} with a cycle of one to train the network for 20 epochs.

\textbf{ILSVRC2012}: We conduct large-scale image classification experiments using the ImageNet ILSVRC2012 dataset~\cite{ILSVRC15}. The evaluation is conducted on the images of the ILSVRC2012 validation set. We use ResNet-18~\cite{he2016deep} as the baseline model. We use SGD~\cite{bottou2010large} with a mini-batch size of 256 to train the network. The initial learning rate is set as 0.1 and weight decay of 0.0001 with a momentum of 0.9. The learning rate is divided by 10 at 30 and 60 epochs. The training is terminated after 90 epochs.

\subsection{Fine-grained image classification}
We follow the training pipeline from~\cite{sun2018multiECCV} to choose ResNet-50 and ResNet-101 as the baseline models. The input images are resized to $448\times448$ for both training and testing and we apply standard augmentation for training data, \ie mirror, and random cropping. The SGD~\cite{bottou2010large} is used to optimize the networks. The learning rate is decayed by 0.1 after 30 and 60 epochs.

\subsection{Multi-class image classification}
We use ResNet-18 with the Multi-Label-Soft-Margin loss as our baseline model. Cyclic cosine annealing~\cite{huang2017snapshot} with the cycle of 1 is used to speed up the training. The total number of training epochs is 20.

% \subsection{Training DeepLab for }
% During training, feature maps are updated and gradients are used to compute the weights for generating attention. According to Fig. 4, the inner layer attends to only local parts of the object, whereas the last layer attends to the object more fully, which results in a better prior for segmentation.
% %Inner layers attend to high frequency content and not the entire target region, resulting in an inaccurate prior for segmentation, so we use only the last layer.
% %Inner layers attend to low-level visual patterns, where attention maps are too noisy to generate reasonable segmentation masks.
% We train Deeplab\footnote{https://github.com/tensorflow/models/tree/master/research/deeplab} in the same way as SEC ~\cite{} is trained in ~\cite{}, using attention maps as weak localization cues.

\section{Multi-label image classification results}
For PASCAL VOC 2012, besides the mean Average Precision (mAP) shown in the main paper, we also provide the results for each category in Table~\ref{tb:pascal}. We notice that ResNet-18 guided by our ICASC$_{\mathcal{A}_{ch}}$ supervision gives the best performance in most of the categories, resulting in the best overall mAP score. When using Grad-CAM~\cite{gradcam} as the attention guidance, the ICASC$_{Grad-CAM}$ also outperforms the baseline method ResNet-18, which further validates the effectiveness of our proposed attention-driven learning framework ICASC.
\begin{table}
\renewcommand\thetable{2}
\centering
\small{
\scalebox{1.0}{
\begin{tabular}{@{\hspace{0mm}}c@{\hspace{2mm}}c@{\hspace{3mm}}c@{\hspace{3mm}}c@{\hspace{0mm}}}
\toprule
class\textbackslash method &ResNet18 & + ICASC$_{Grad-CAM}$ & + ICASC$_{\mathcal{A}_{ch}}$ \\\hline
	aeroplane & 95.16  & 96.33 & \textbf{96.85} \\
	bicycle & 76.18 & 80.82 & \textbf{82.41}  \\
	bird & 92.92 &94.69 & \textbf{95.17}  \\
	boat & 84.82  &87.91 &\textbf{89.13} \\
	bottle & 53.32 &60.66 &\textbf{61.07}  \\
	bus & 89.81  &91.73 &\textbf{92.25} \\
	car & 77.41  &80.11 &\textbf{81.74}  \\
	cat & 93.91  &95.63 &\textbf{96.28}  \\
	chair & 68.53  &73.04 &\textbf{73.69}  \\
	cow & 57.41  &67.85 &\textbf{71.12}  \\
	diningtable & 67.35  &73.07 & \textbf{73.64}  \\
	dog & 88.18  &91.70 &\textbf{92.62}  \\
	horse & 73.58  &80.89 &\textbf{84.21}  \\
	motorbike & 82.36  &86.09 &\textbf{87.51}  \\
	person & 95.69  &96.22 &\textbf{96.44}  \\
	pottedplant & 46.38  &56.27 & \textbf{57.75}  \\
	sheep & 78.79  &84.93 &\textbf{86.15}  \\
	sofa & 54.83  &64.00 &\textbf{64.63} \\
	train & 92.05  &95.05 &\textbf{95.56}  \\
	tvmonitor & 80.07  &\textbf{85.32} & 85.24 \\\hline
	mAP & 77.44  & 82.12 &\textbf{83.17} \\ \bottomrule
\end{tabular}
} % added scale box
\vspace{0.5mm}
\caption{Categorical and mean Average Precision (mAP) (\%) for our PASCAL VOC 2012 image classification experiment. The highest scoring entry in each row is shown in bold.}
\vspace{-2.5mm}
\label{tb:pascal}
}
\vspace{-1.5mm}
\end{table}

\section{Additional qualitative results}
\begin{figure*}
\renewcommand\thefigure{1}
    \centering
    \includegraphics[width=\textwidth]{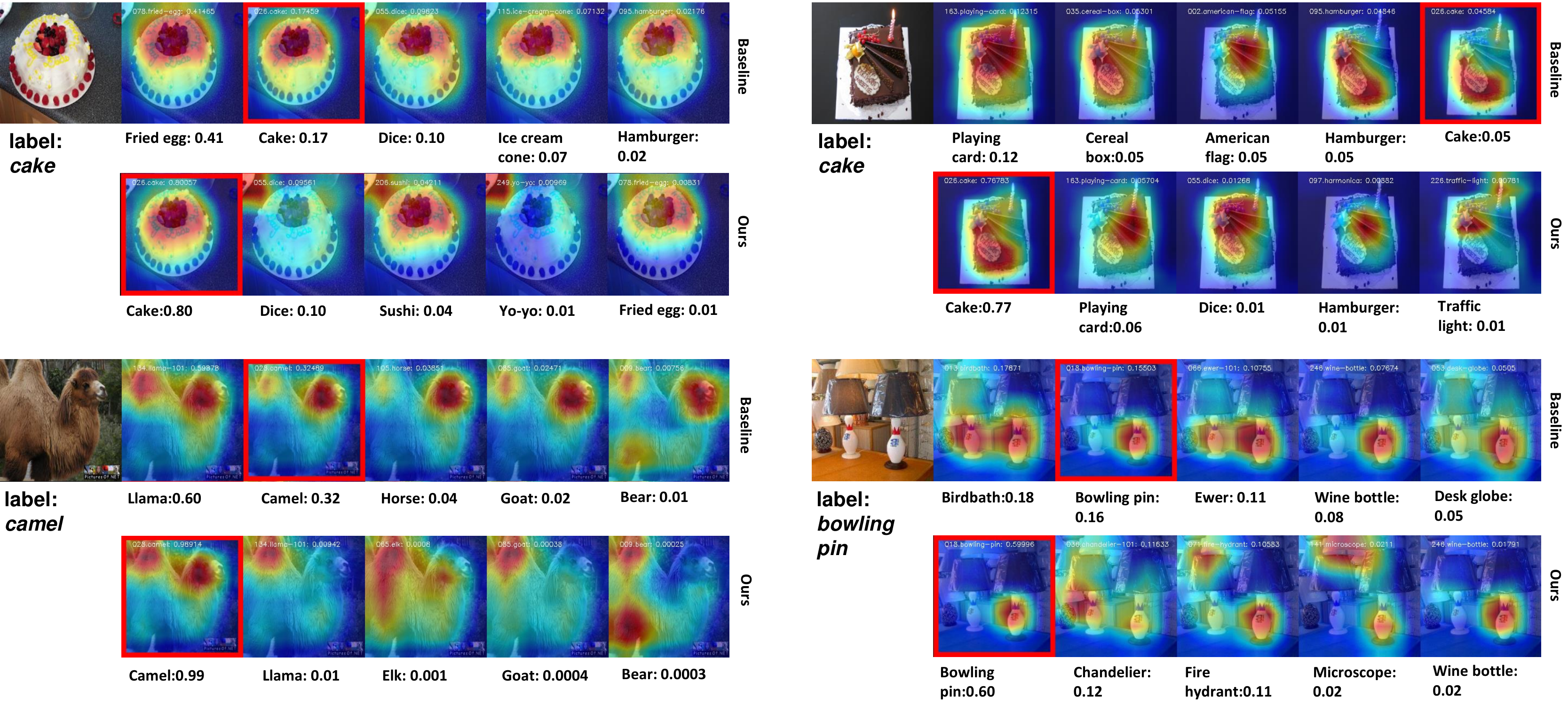}
    \caption{Improvements in top-1 predictions with our method (ResNet-18 + ICASC$_{\mathcal{A}_{ch}}$) when compared to the baseline (ResNet-18). Top row: ResNet-18; bottom row: ResNet-18 + ICASC$_{\mathcal{A}_{ch}}$.}
    \label{fig:exp1}
\end{figure*}

\begin{figure*}
\renewcommand\thefigure{2}
    \centering
    \includegraphics[width=\textwidth]{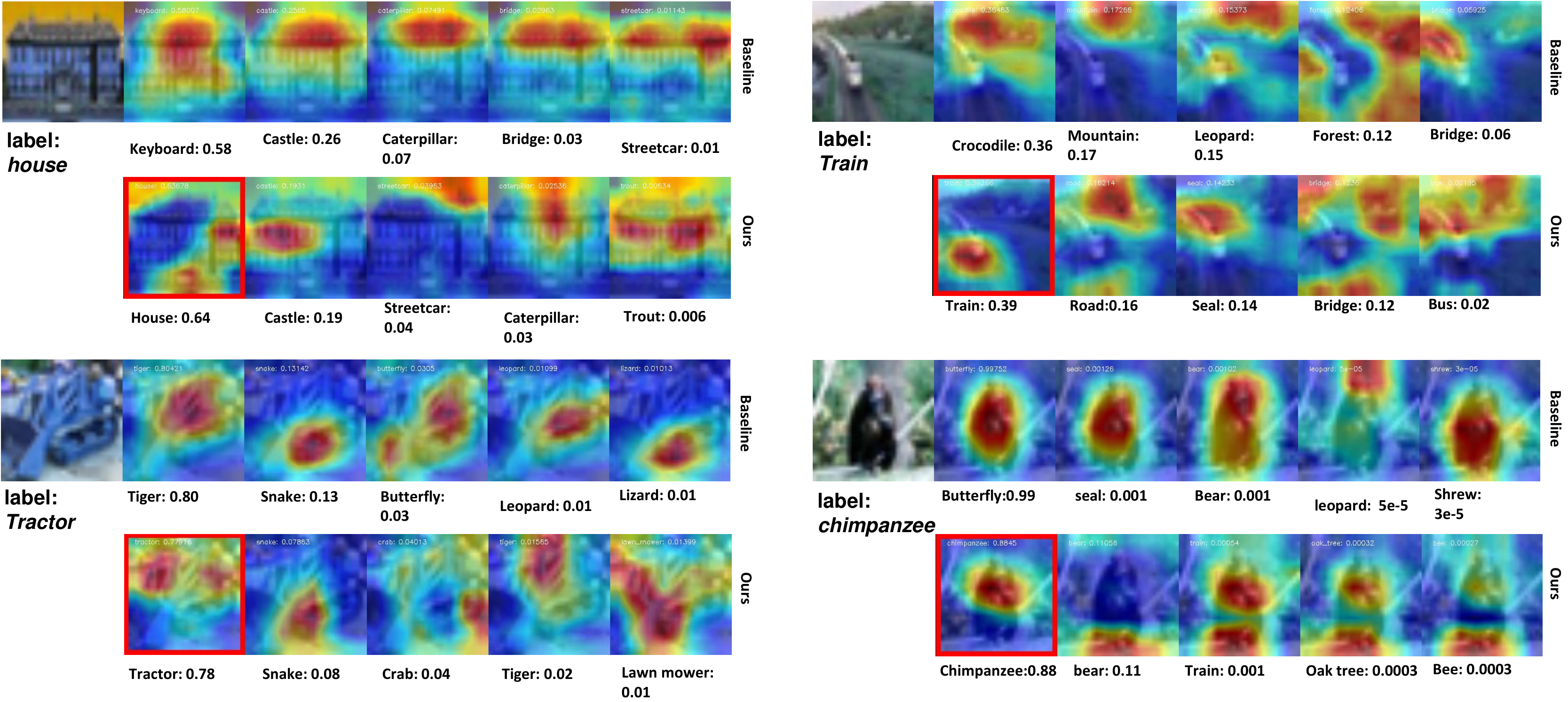}
    \caption{Improvements in top-1 predictions with our method (ResNet-110 + ICASC$_{\mathcal{A}_{ch}}$) when compared to the baseline (ResNet-110). Top row: ResNet-110; bottom row: ResNet-110 + ICASC$_{\mathcal{A}_{ch}}$.}
    \label{fig:exp4}
\end{figure*}

We show additional qualitative results for our proposed method in Figure~\ref{fig:exp1} %~\ref{fig:exp2},~\ref{fig:exp3},
and Figure~\ref{fig:exp4}. %\knote{If possible, for each figure, I suggest that we show the top-5 labels and the corresponding classification scores just like Figure 10 in the main paper. In addition, please specify the exact methods in each figure (I assume it's ResNet-110 (top row) and ResNet-110 + ICASC$_{\mathcal{A}_{ch}}$ (bottom row)).}
Each figure shows four examples, where for each example, we show the input image and the ground-truth class in the first column, the top-5 categorical attention maps for the baseline in the top row of the adjacent columns, and those with our approach in the bottom row. In Figure~\ref{fig:exp1}, %and~\ref{fig:exp2},
where the images are in high resolution, the baseline method is ResNet-18 and our method is ResNet-18 + ICASC$_{\mathcal{A}_{ch}}$. In %Figure~\ref{fig:exp3}
Figure~\ref{fig:exp4}, the baseline method is ResNet-110 and our method is ResNet-110 + ICASC$_{\mathcal{A}_{ch}}$. In all the figures, the ground-truth class attention map is marked using a red bounding box. There will be no marked attention map if the ground-truth class is not in the top-5 predictions. These figures show that our discriminative attention achieves better attention separability, with our model attending to regions that tell different categories apart. On the other hand, we observe visual confusion with the baseline, with high responses in the attention maps located at similar spatial locations among different categories.

As can be seen from these figures, since discriminative attention is our principled learning objective, attention responses given by our method across the top-5 categories are more separable than those from the baseline method, and our trained model is able to attend to semantically discriminative parts of the ground-truth objects, resulting in the better classification results. For example, in the top left ``cake" example in Figure~\ref{fig:exp1}, for both ``cake" and ``fried egg," the baseline method attends to the central areas, containing the fruits and the cream around, which leads to visual confusion and misclassification of the image as ``fried egg," whereas our method attends to the central part (fruits and cream) for ``cake" and the right part (cream) for ``fried egg," classifying the image as ``cake" correctly.
% Figure~\ref{fig:exp2},~\ref{fig:exp3}, and~\ref{fig:exp4} also validate the fact that our attention-driven learning process ICASC generates more separable attention maps than the baseline and results in better classification performance, compared to the baseline method.
Additionally, in Figure~\ref{fig:exp4}, our method brings the ground-truth class to the top-1 which is out of top-5 predictions in the baseline method.

%target category a distinguishing the target categories. While, there are visual confusion in base model where attentions are located at the same regions for different classes, resulting incorrect top-1 prediction.

% \begin{figure*}
% %%%\renewcommand\thefigure{12}
%     \centering
%     \includegraphics[width=\textwidth]{imgs/exp1.pdf}
%     \caption{Improvements in top-1 predictions with our method (ResNet-18 + ICASC$_{\mathcal{A}_{ch}}$) when compared to the baseline (ResNet-18). Top row: ResNet-18; bottom row: ResNet-18 + ICASC$_{\mathcal{A}_{ch}}$.}
%     \label{fig:exp2}
% \end{figure*}

% \begin{figure*}
% %%\renewcommand\thefigure{13}
%     \centering
%     \includegraphics[width=\textwidth]{imgs/exp3.pdf}
%     \caption{Improvements in top-1 predictions with our method (ResNet-110 + ICASC$_{\mathcal{A}_{ch}}$) when compared to the baseline (ResNet-110). Top row: ResNet-110; bottom row: ResNet-110 + ICASC$_{\mathcal{A}_{ch}}$.}
%     \label{fig:exp3}
% \end{figure*}

% \begin{table*}[h]
% %\renewcommand\thetable{12}
% \centering
% \small{
% \begin{tabular}{c@{\hspace{3mm}}c@{\hspace{3mm}}c@{\hspace{3mm}}c@{\hspace{3mm}}c}
% \toprule
% dataset &CIFAR-100
% &Caltech-256
% &ImageNet
% &PASCAL VOC 2012\\
% \midrule
% \# classes &100 &256 &1000 &20\\
% image size &32$\times$32 &299$\times$299 &224$\times$224 &299$\times$299\\
% \# images &60000 &30607 &$\sim$1.3M  &11540\\
% \midrule
% \# training images &50000 &7680(N=30)/15360(N=60) &1.2M &5717\\
% \# testing images &10000 &6400(N=25) &50000  &5823\\
% %data split &specified \red{\cite{}} &random %&specified \red{\cite{}} &specified %\red{\cite{}} &specified \red{\cite{}}\\
% training batch size &128 &32 &256 &32\\
% %weight decay &$0.0005$ &$10^{-3}$ &$10^{-4}$ &$10^{-3}$\\
% %momentum &0.9 &0.9 &0.9 &0.9 &0.9\\
% initial learning rate(lr) &0.1 &0.01 &0.1 &0.01\\
% %lr adjustment &step_decay &0.01 &0.1 %&$10^{-3}$ &0.01\\
% \# training epochs &160 &20 &90  &20\\
% evaluation metric &Top-1 Accuracy &Top-1 Accuracy &Top-1 Accuracy &mean Average Precision\\
% %reference of above setting &~\cite{he2016deep} &- &~\cite{huang2017densely} &\cite{liu2016fully,sun2018multiECCV} &-\\
% \bottomrule
% \end{tabular}
% \vspace{.5em}
% \caption{The details of the dateset and training parameters. The references of training recipes are listed at the bottom}
% \label{table:dataset}
% }
% \end{table*}

\end{appendices}
\end{document}